\documentclass[fleqn,10pt]{utils/wlscirep}
\usepackage[utf8]{inputenc}
\usepackage[T1]{fontenc}
\usepackage{lineno}
\usepackage{graphicx}
\usepackage{enumitem} 
\usepackage{subcaption}
\usepackage{cleveref}
\usepackage{tabularx}
\usepackage{multirow}
\title{On the Readiness of Scientific Data Papers for a Fair and Transparent Use in Machine Learning}

\author[1,4]{Joan Giner-Miguelez\thanks{Corresponding author: joan.giner@bsc.es}}
\author[1]{Abel Gómez}
\author[2,3]{Jordi Cabot}
\affil[1]{Internet Interdisciplinary Institute (IN3), Universitat Oberta de Catalunya (UOC), Barcelona, Spain}

\affil[2]{Luxembourg Institute of Science and Technology, Esch-sur-Alzette, Luxembourg}
\affil[3]{University of Luxembourg, Esch-sur-Alzette, Luxembourg}
\affil[4]{Barcelona Supercomputing Center, Plaça Eusebi Güell, 1-3, Barcelona, Spain}

\begin{abstract}

To ensure the fairness and trustworthiness of machine learning (ML) systems, recent legislative initiatives and relevant research in the ML community have pointed out the need to document the data used to train ML models. Besides, data-sharing practices in many scientific domains have evolved in recent years for reproducibility purposes. In this sense, academic institutions' adoption of these practices has encouraged researchers to publish their data and technical documentation in peer-reviewed publications such as data papers. In this study, we analyze how this broader scientific data documentation meets the needs of the ML community and regulatory bodies for its use in ML technologies. We examine a sample of 4041 data papers of different domains, assessing their completeness, coverage of the requested dimensions, and trends in recent years. We focus on the most and least documented dimensions and compare the results with those of an  ML-focused venue (NeurIPS D\&B track) publishing papers describing datasets. As a result, we propose a set of recommendation guidelines for data creators and scientific data publishers to increase their data's preparedness for its transparent and fairer use in ML technologies.

\end{abstract}
\begin{document}

\flushbottom
\maketitle

\thispagestyle{empty}

%\noindent Please note: Abbreviations should be introduced at the first mention in the main text – no abbreviations lists or tables should be included. Structure of the main text is provided below.
\vspace{2mm}
\section{Introduction}

The growing influence of machine learning (ML) technologies in our society raises scientific and political concerns about the potential harms they may cause. Datasets are critical in these systems, and recent research has highlighted them as one of the root causes of unexpected and harmful consequences in ML applications. Recent studies, for example, observed gender-biased classifiers for computer-aided diagnosis due to imbalances in training data \cite{genderimbalance, abbasi2020risk}, or ML models for pneumonia detection failing to generalize to other hospitals due to specific conditions during the collection of the images used to train them \cite{zech2018variable}. This situation has prompted the interest of regulatory agencies and the ML community in general in developing data best practices, such as building proper dataset documentation. Public regulatory initiatives, such as the European AI Act and the AI Right of Bills, as well as relevant scientific works \cite{datasheets, ginerDSL}, have proposed general guidelines for developing standard dataset documentation. Such proposals identify a number of dimensions, such as the dataset's provenance or potential social issues, that could influence how the dataset is used and the quality and generalization of the ML models trained with it.

Interest in data-sharing practices was already widespread in many scientific fields, determined, among others, by the need to reproduce scientific experiments \cite{tedersoo2021data}. The adoption of FAIR principles \cite{wilkinson2016fair} and data management plans \cite{rolando2015data}  by research institutions has encouraged researchers to publish their data with technical and scientific documentation, such as data papers. Recent research has focused on how to improve data-sharing practices through this type of documents, working on specific aspects such as the peer-review process for data papers \cite{mayernik2015peer}, data citation practices \cite{silvello2018theory}, data publishers' guidelines \cite{kim2020analysis}, and data reuse \cite{faniel2019context}. 

The use of scientific data published through data papers for the training of ML models is making good data practices even more important. However, research has yet to be conducted on the suitability of these data-sharing practices of the broader scientific community and the data dimensions required by the ML community. In that sense, this paper aims to study the presence (and evolution) of such dimensions in current data papers published by the scientific community. To do so, we analyzed 4041 open access data papers published in two interdisciplinary data journals: Nature's \emph{Scientific Data} and Elsevier's \emph{Data in Brief}. We examined the full manuscripts regarding the demanded dimensions, which are detailed in Section~\ref{sec:back}, and we compared the results with an ML-focused venue (NeurIPS Datasets and Benchmarks track) publishing papers
describing datasets. The analysis was aided by an ML pipeline that included a Large Language Model (LLM) at its core, designed to extract specific dimensions from the manuscripts. 

Using the extracted data, we aimed to answer the following research question: 

\begin{itemize}

\item RQ1: To what extent are data papers documented with the dimension demanded by the recent ML documentation frameworks?

With this question, we aim to see to what extent the data documentation practices outside the ML field already answer the needs stated by the recent ML documentation frameworks.

\item RQ2: How have documentation practices evolved in the last few years regarding the dimension the ML community demands?

With this question, we aim to see how the recent application of documentation frameworks in the ML community has influenced broader data publishing venues. 

\end{itemize}

%Even though these scientific data publications are becoming high-quality and well-documented data sources for the ML community, research has yet to be conducted on the intersection between them and the needs of the ML community. In this study, we bridge the gap between them by analyzing the presence and trends of demanded dimensions of the ML community in data papers.

As we detail in our results---cf. Section \ref{sec:res}---we noticed a significant contrast in the presence of the required dimensions. Those already explicitly requested in the publisher submission guidelines, such as the data recommended uses or a general description of the collection process, are consistently present. In contrast, details about the data generalization limits and its social concerns,  the profiles of the team collecting and annotating the data, and the data maintenance policies need to be much better documented if we want to ensure ML models trained on trustworthy and reliable data. To help with this, in Section \ref{sec:dis}, we propose a set of recommendation guidelines for data creators and scientific data publishers to increase the presence of the least documented dimensions as a way to improve their data for its transparent and fair use in machine learning. The ML extraction pipeline and the entire dataset used in the analysis are openly available \cite{zenodoRepo}.

\section{Background}
\label{sec:back}

The need for proper documentation for datasets in the ML field is well defined in the well-known publication \emph{Datasheets for Datasets}\cite{datasheets}, a work inspired by datasheets in the electronics industry. In this work (among others \cite{mcmillan2021reusable, bender-friedman-2018-data, holland2020dataset, zech2018variable, sasha}), the authors identify data aspects for each phase of the data creation process---design, gathering, and annotations---that could affect how the dataset should be used or the quality of ML models trained. Table 1 summarizes the dimensions covered by recent documentation frameworks, representing a common ground between them. This list of dimensions is based on the previous author's work  \cite{ginerDSL} and the Responsible AI extension of Croissant  \cite{akhtar2024croissant}, a metadata language for describing ML datasets promoted by the MLCommons consortium.%\footnote{Croissant taskforce homepage: \url{https://mlcommons.org/working-groups/data/croissant/}}.

\begin{table}[b!]
    \centering
    \caption{Dimensions analyzed from the sample of scientific data papers \\}
    \label{dimensions}
    \begin{tabular}{lll}

       \hline
        \textbf{Dimensions} & \textbf{Subdimensions} & \textbf{Target explanation} \\ \hline   
        {\textbf{Uses}} & Recommended uses &  Recommended uses and gaps the dataset intends to fill \\ 
        \textbf{} & Generalization limits  & Non-recommended uses and data generalization limits  \\ 
        \textbf{} & Social concerns  & If represent people: biases, sensitivity and privacy issues  \\ 
        
        \textbf{} & Maintenance Policies  & Maintainers \& policies (erratum, updates, deprecation) \\  
        
        \textbf{} & Tested using an ML approach  & Models and metrics the dataset have been tested on  \\  \hline

        {\textbf{Collection}} & Description  & Description of the process and its categorization  \\ 
        \textbf{} & Profile of the collection team  & Profile of the gathering team \\ 
        \textbf{} & Profile of the collection target  & If represent people, their demographic profile \\ 
        \textbf{} & Speech context in language datasets  & If language datasets, contextual information of the spoken language \\ 
        \textbf{} & Collection sources \& infrastructure  & The source of the data and the infrastructure used to collect it \\ 
  
        \hline
      
        {\textbf{Annotation}} & Description  & Description of the process and its categorization \\ 
        \textbf{} & Profile of the annotator team  & Profile of annotation the team  \\ 
        \textbf{} & Annotation infrastructure  & The tools used to annotate the data \\ 
        \textbf{} & Annotation validation  & Validation methods applied over the annotations \\ \hline

        \hline

    \end{tabular}

\end{table}

The \emph{Uses} dimensions of the data refer to how this data should or should not be used and its potential social effects. In terms of \emph{Recommended uses}, \emph{Datasheets for Datasets}, among other works\cite{mcmillan2021reusable, bender-friedman-2018-data, holland2020dataset}, propose to document the purposes and gaps the dataset intends to fill. Also, they demand rationales about the \emph{Generalization limits} of the data and the potential \emph{Social concerns} if the data represents people. For instance, data gathered from patients of a single hospital may have generalization limits for various reasons. The most obvious is data distribution, where, for instance, a cohort of patients is not representative of the whole population \cite{zech2018variable}. Also, some of the ML community works \cite{sasha} express concerns about the \emph{Maintenance} of the data. Setting an erratum to inform potential errors, an updating timeframe, or deprecation policies are remarked as useful practices while releasing data.

Data provenance is also the focus of these works. Documenting specific details about the \emph{Collection} and \emph{Annotation} process is stated as a need to evaluate properly the suitability of the data for specific applications. For instance, recent studies \cite{zech2018variable} have shown that the use of specific devices and practices during the collection of chest lung cancer images could raise a biases in data, leading to spurious correlations \cite{garbage} of the ML model, and providing incorrect accuracy metrics. These examples illustrate the usefulness of knowing information about the \emph{Sources \& Infrastructure}, such as the scanners used to get MRI images, and other contextual attributes of the data. As another example, the annotation \emph{Infrastructure} used, such as a labeling software, and the \emph{Validation} methods applied over the labels could reduce the potential bias of the data and could give us clues about the data generalization limits. In that sense, a data user may provide more confidence to a record of data labeled by a group of people together with the inter-annotator agreement between them than labeled by a single one.

Also, there is a demand to characterize the people who have participated in the provenance steps of the data, as they add subjectivity to it \cite{healthcare}. For instance, a dataset curated by a team of experts in the field or curated by a crowdsourcing service, such as \emph{Amazon Mechanical Turk}, could not assign the same values. Therefore standalone data without information about the team involved in its creation may not be useful to evaluate the suitability of the data for an application. As creating data involves more people every time, there is a need to go beyond ``the authors'' of the dataset and \emph{Profile the collection team} and \emph{the annotation team} of the data. Beyond those who actively created the data, recent works \cite{diaz2022crowdworksheets} also stress the need to profile passive participants of the dataset creation. For instance, if the data is collected by or represents people, a profile of these people should also be required (\emph{Profile of the collection target}). This practise is widely integrated in health data where the profile of the cohort of patients is highly relevant. Finally, for natural language data, which has gained relevance to the recent emergence of large language models, 
recent works \cite{bender-friedman-2018-data} also stress the need to document contextual information about the speech (\emph{Speech Context}) such as the specific dialect or the modality of the text. For instance, a natural language dataset gathered from Australian speakers may reduce the accuracy of models trained to support users in the United States due to different accents and language communication styles.

\begin{figure}[b]
    \centering
    \includegraphics[width=7cm]{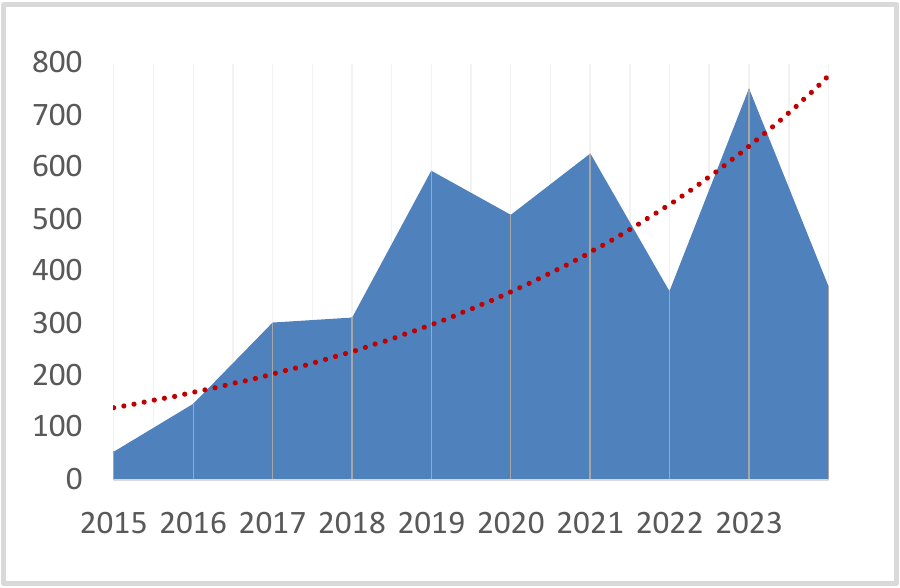}
    \caption{\label{fig:paperperyear}  Number of data papers published between 2015 and 2023 evaluated in the sample. 2023 has been evaluated until June.}
\end{figure}

\section{Results}
\label{sec:res}
This section reports on the results of our analysis. We first provide some information on the representativeness of the sample, describing the size and diversity of the data papers, and second, we report on the presence or absence of the information dimensions that characterize a good dataset for ML training purposes. In this report, we provide an overall picture of the presence of the dimensions, a temporal evolution of such data to see whether the situation is improving, and a comparison between the two analyzed data journals. Then, we present the results of analyzing the data papers of an ML-focused venue (the NeurIPS track on datasets and benchmarks) to draw a picture of the current documentation practices in the ML community and to compare it with the results obtained.

\begin{table}[t]
   
    \caption{\emph{Uses}: Extract of the 10 most popular topics in data uses of the analyzed data papers \\}
    \label{topics}
    \small
    \begin{tabularx}{\textwidth}{l p{3cm} p{9cm}}
    
       \hline
         \multirow{2}{4cm}{\textbf{Topic}} &  \multirow{2}{3cm}{ \textbf{Related data papers \\ (Sample's percentage)}} & \multirow{2}{9cm}{\textbf{Example}}  \\ \\
            \\  \hline 
        \multirow{2}{4.5cm}{RNA sequencing for gene analysis} & \multirow{2}{1.4cm}{371 (9,1\%)} &  "Quality control and data analysis in RNA-seq experiments, including quality analysis of raw sequencing data..." \cite{emon2023mechanosensitive}
        \\ \hline 
        \multirow{2}{4cm}{Medical Imaging} & 
        \multirow{2}{{1.4cm}}{330 (8,1\%)} &  "MRI image processing procedures and constructing breast models for studying the effects of chemotherapy on breast cancer patients." \cite{ALDHABYANI2020104863}
         \\  \hline 
        \multirow{2}{4.5cm}{Chemistry and physics} & \multirow{2}{{1.4cm}}{232 (5,7\%)} &  "Soil chemistry analysis, soil classification, and land use analysis in New Zealand's major agricultural regions." \cite{shen2018data}
        \\   \hline 
        \multirow{2}{4.5cm}{Material properties} & \multirow{2}{{1.4cm}}{202 (4,9\%)} &  "Analyze and study solubility data, perform high throughput DFT calculations..." \cite{xing2021optical}
          \\   \hline 
        \multirow{2}{4.5cm}{Climate and hydrological analysis} & \multirow{2}{{1.4cm}}{189 (4,6\%)} &  "Preprocessing of functional MRI data, including motion correction, slice-timing correction, non-brain removal..." \cite{aerts2022pre}
          \\   \hline 
        \multirow{2}{4.5cm}{Biological conservation} & \multirow{2}{{1.4cm}}{160 (3,9\%)} &  To study the biodiversity and distribution of fish species, and assess the conservation status of fish species." \cite{rodeles2016iberian}
          \\   \hline 
        \multirow{2}{4.5cm}{Agricultura and cropland} & \multirow{2}{{1.4cm}}{87 (2,1\%)} &  "Optimizing agricultural production strategies and water resources management, as well as validating crop yield..." \cite{cheng2022high}
          \\   \hline 
        \multirow{2}{4.5cm}{Drug development} & \multirow{2}{{1.4cm}}{72 (1,8\%)} & "Analyze drug-indication pairs and to study the relationship between drugs and diseases..."\cite{brown2017standard} 
          \\   \hline 
        \multirow{2}{4.5cm}{Geological and Seismic analysis} & \multirow{2}{{1.4cm}}{54 (1,4\%)} & reconstructions of paleoclimate data and to analyze the variables of interest for.. \cite{steiger2018reconstruction}
                  \\   \hline 
        \multirow{2}{4.5cm}{Human movement recognition} & \multirow{2}{{1.4cm}}{53 (1,3\%)} &   analyzing locomotor trials and standing trials in healthy subjects and persons with motor disturbances... \cite{lencioni2019human}
          \\   \hline

\end{tabularx}

\end{table}

\begin{figure}[!b]
   \begin{minipage}{1\textwidth}
    \centering
    \includegraphics[width=0.7\linewidth]{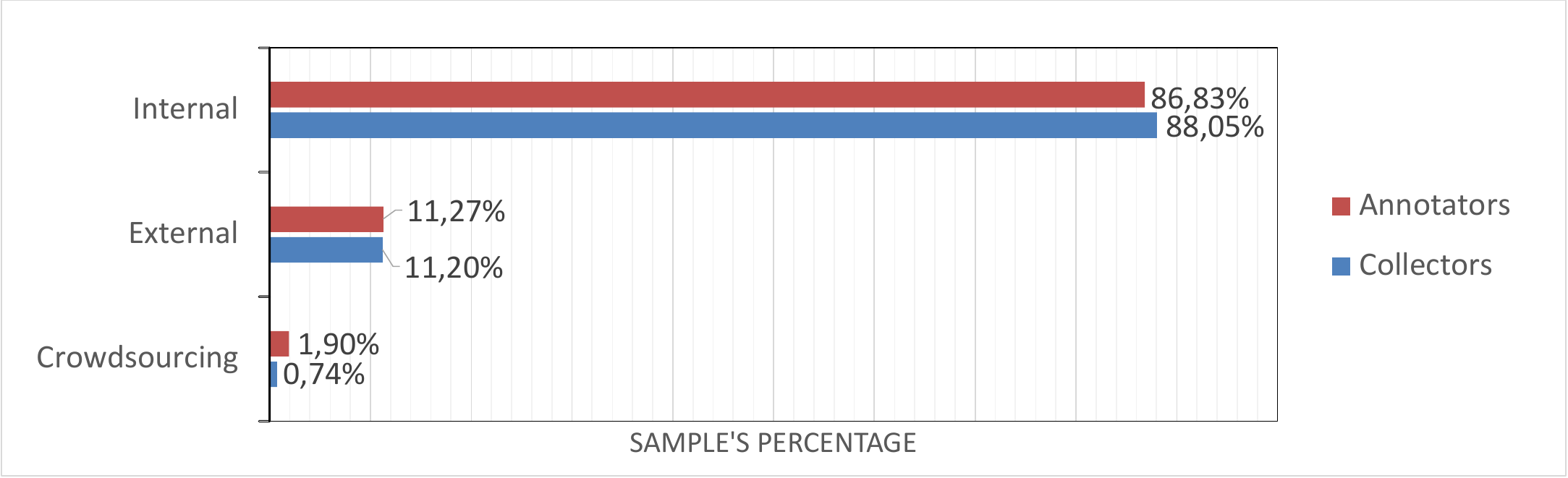}
    \caption{\label{fig:teams}  Diversity in the collection and annotation teams type of the analyzed data papers}
    \vspace{2mm}
\end{minipage}
   \begin{minipage}{0.48\textwidth}
     \centering
     \includegraphics[width=0.85\linewidth]{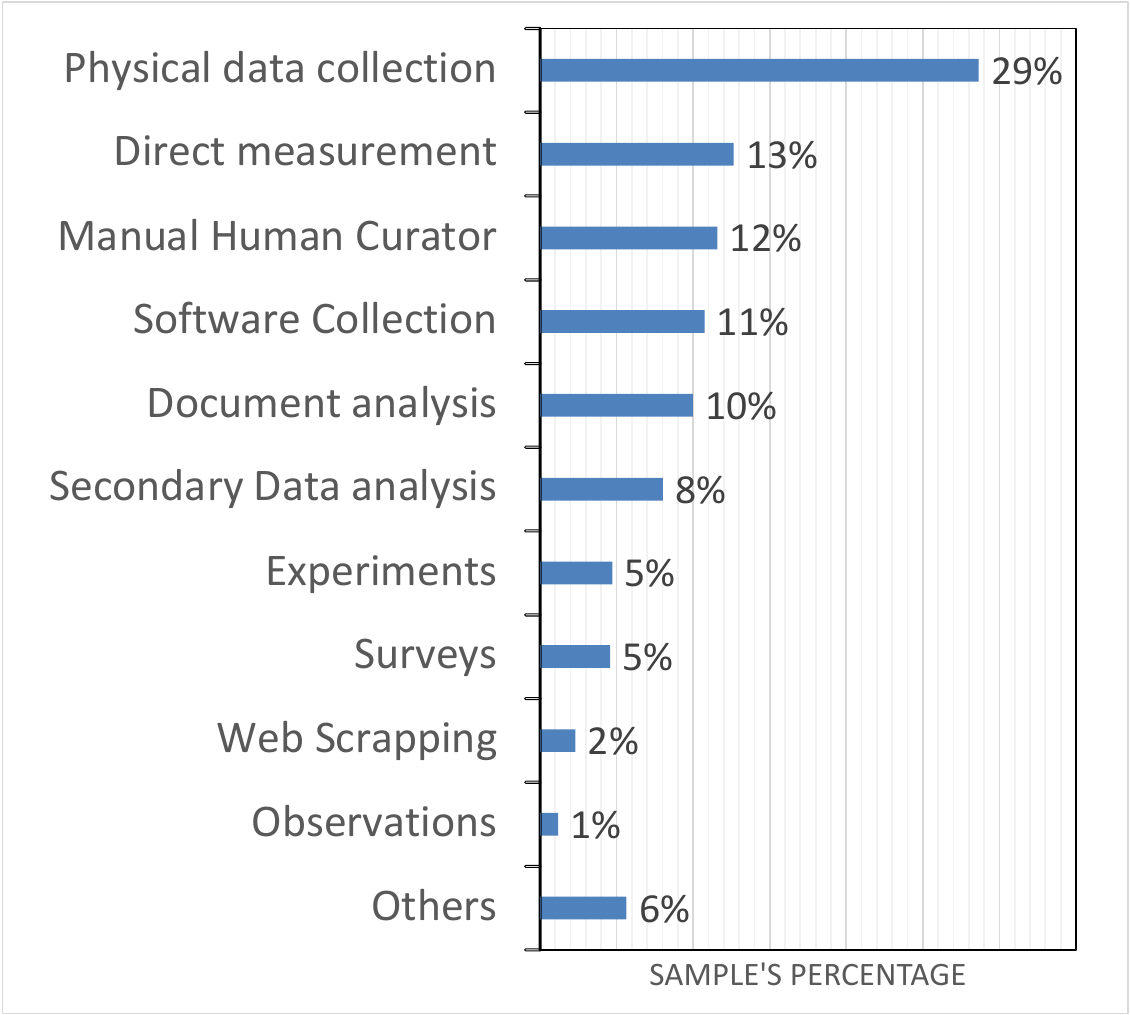}
     \caption{\emph{Collection}: Diversity of types of collection processes}\label{Fig:Collect}
   \end{minipage}\hfill
   \begin{minipage}{0.48\textwidth}
     \centering
     \includegraphics[width=0.85\linewidth]{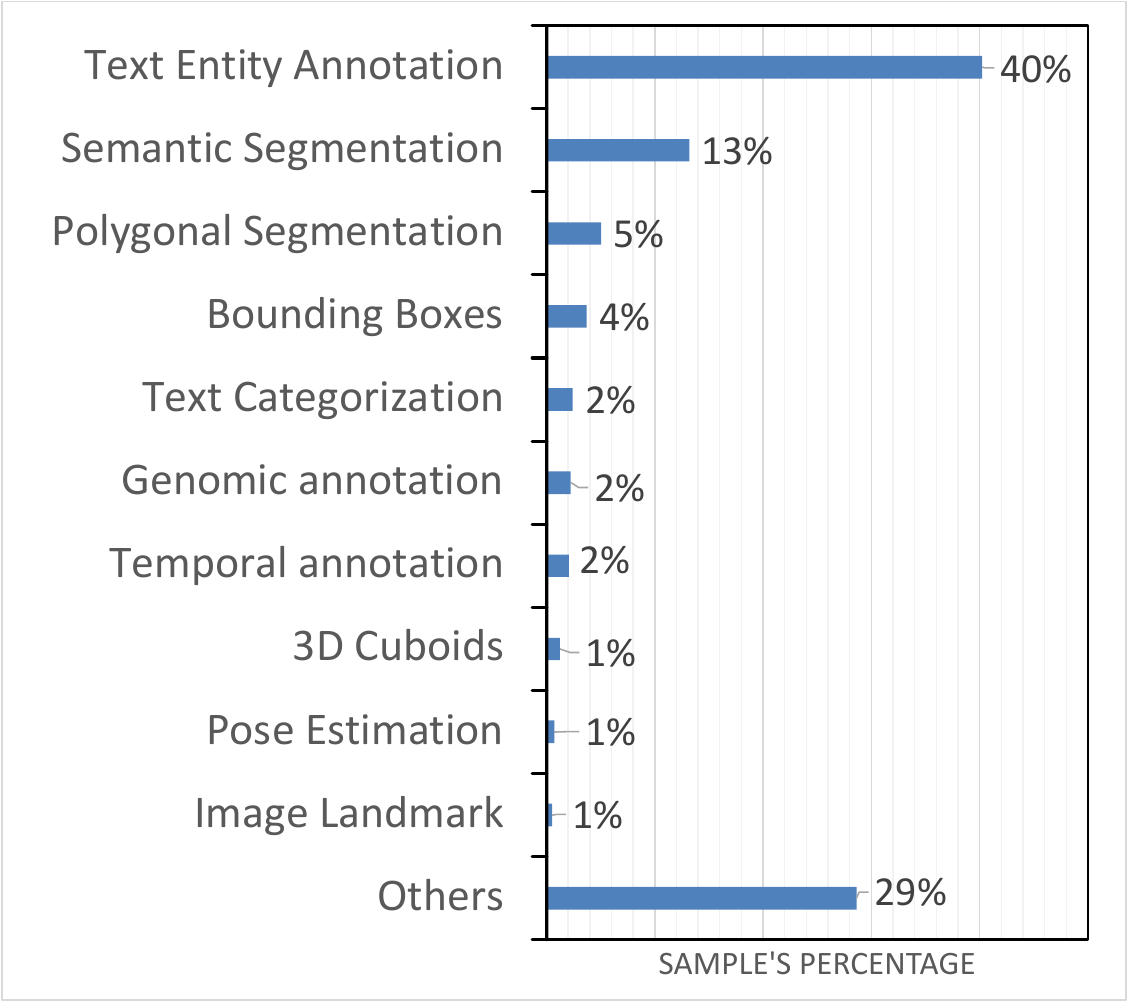}
     \caption{\emph{Annotation}: Diversity of types of annotation processes}\label{Fig:Annotate}
   \end{minipage}
\end{figure}

\subsection{Size and diversity of the sample}
The analyzed sample comprises 4041 data papers extracted from two interdisciplinary scientific journals. In Figure~\ref{fig:paperperyear}, we can see an apparent growing trend in the number of papers published between 2015 and 2023, with 2021 being an exception due to the effect of COVID-19 on general scientific production. Note also that the absolute value for 2023 is smaller as we only collected data until June of 2023. The 16,5\% of the sample's data papers are related to people, and this is the subsample we used to calculate the presence of the \emph{Social concerns} and the \emph{Profile of the collection targets} dimensions in the next section as these metrics are only relevant for people-related datasets.

As an indication of the diversity of our dataset, Table~\ref{topics} shows the 10 most popular topics of the uses of the data in the analyzed sample, showing the sample diversity. The topics related to the \emph{Conservation of biodiversity}, \emph{RNA sequencing for gene analysis}, and \emph{Chemistry and Physics} are the most common ones. Other topics, such as \emph{Breast Cancer} and \emph{Magnetic Resonance Imaging} (MRI), are very interrelated, where most of the datasets around Breast Cancer are composed of MRI images. On the other hand, topics such as COVID-19 have appeared quickly in 2021 and are decreasing over time, as they are strongly correlated with a temporal event such as the coronavirus pandemic. The full results of the topic analysis can be found in our open repository \cite{zenodoRepo}.

Our sample shows diversity in terms of the \emph{annotation} and \emph{collection} processes types as well. For instance, \mbox{Figure \ref{Fig:Collect}} depicts the distribution of the most common data collection types with \emph{Physical data collection, Direct measurements, Document analysis, Manual human curator}, and \emph{Software collection} (all above 10\%) being the most common ones. %It could be noted that only 6\% of the sample has been classified as an \emph{Others}. 
Moreover, Figure \ref{fig:teams} shows that nearly 88,05\% of the teams who participated in data collection were internal teams (authors of the study or close collaborators), 11,20\% were external teams, and only 0,74\% used crowdsourcing practices to collect the data. Regarding the annotation process, only 42,28\% of the analyzed data papers employ an annotation process involving humans. Figure \ref{Fig:Annotate} shows that the most common type of annotation is \emph{Text Entity Annotation}, followed by \emph{Semantic segmentation}, \emph{Polygonal segmentation} and \emph{Bounding boxes}. Compared to the collection processes, 29\% of the processes were classified as \emph{Others}. The results for team type are similar to the results for the collection process, with a high percentage of external teams (11,27\%) and crowdsourcing teams (1,90\%).

\subsection{Dimensions presence in scientific data documentation}

\begin{figure}
   \begin{minipage}{0.9\textwidth}
    \centering
     \includegraphics[width=1\textwidth]{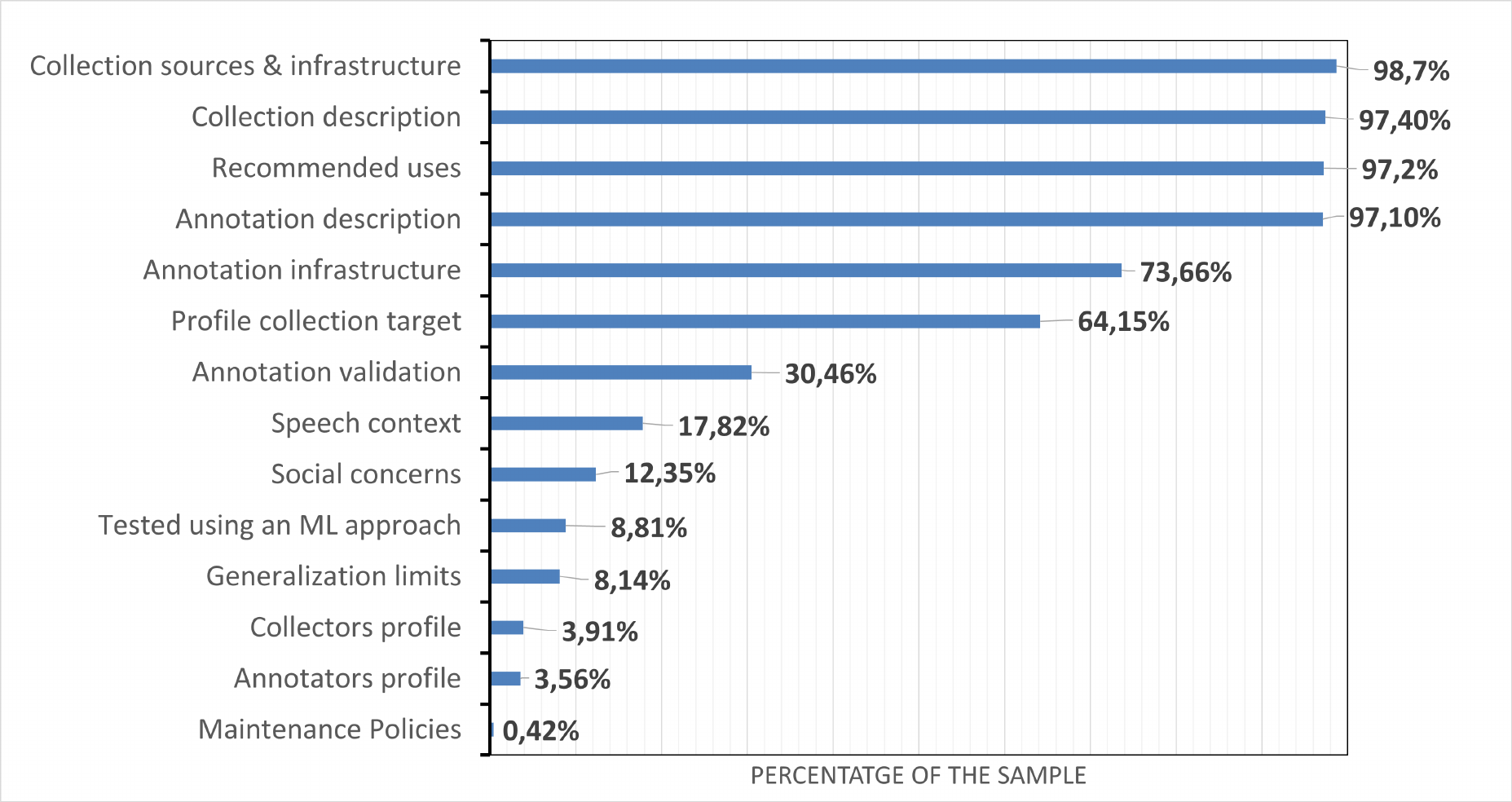}
     \caption{Overall results of informed dimensions. \emph{Social concerns} and \emph{Profile of the collection targets} dimension have been evaluated only on datasets gathered from or describing people (16,5\% of the sample). \emph{Speech context in language datasets} has only been assessed on datasets representing natural language (5,15\% of the sample). \emph{Annotation} dimensions have been assessed only on datasets created through an annotation process (42,28\% of the sample). In these cases, the percentage reflects the occurrence of those dimensions relative to the number of papers that should declare them.}
     \label{fig:years}
  \end{minipage}\hfill
 \vspace{5mm}
\begin{minipage}{0.9\textwidth}
\centering

\includegraphics[width=1\textwidth]{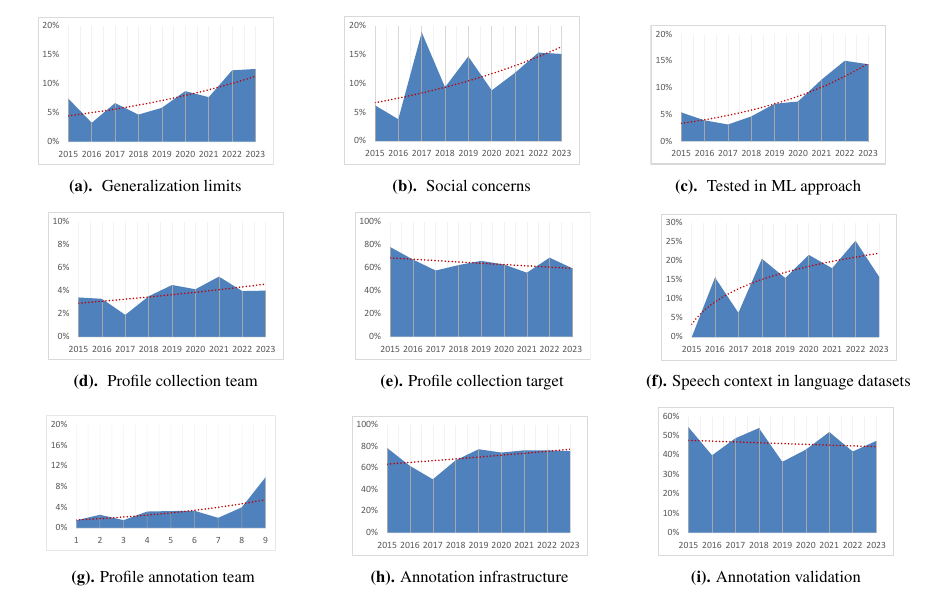}

\caption{Evolution of the least informed dimensions from 2015 to 2023 (June). Figure (b) is evaluated only on datasets gathered from or describing people (16,5\% of the sample). Figure (e) and Figure (f) are evaluated only on datasets collected from or representing people. Figure (f) is evaluated only on natural language datasets. Finally, Figures (g), (h), and (i) are evaluated only on datasets created trough an annotation process.}
\label{fig:tendency}
  \end{minipage}\hfill
  
\end{figure}

% Molt informades
In this section, we go across the dimension presented in Section \ref{sec:back}, analyzing its presence in the  data papers. Figure \ref{fig:years} depicts the overall results. We can see that dimensions such as the \emph{Recommended uses} (97,2\%), \emph{Collection description} (97,4\%), \emph{Collection sources \& infrastructure} (96,1\%), and the \emph{Annotation description} (97,1\%) have a very high presence%\footnote{Due to the extraction method's tolerance, very high percentages can be understood as dimensions informed in all the analyzed data papers.}.
This is a sharp contrast with several other dimensions that are much less documented. For these less-documented dimensions, we depict their evolution between 2015 and 2023 in Figure \ref{fig:tendency}, and in the remainder of this section, we will take a closer look at them. Some of these show an encouraging improvement over the last years but still remain at very low level. Note that our analysis focuses on whether a given metric is present or not in the dataset. We do not penalize data papers if the metric is present but with a low-level detail or incomplete.

\textbf{Generalization limits:} Only  8,14\% of the data papers report on the \emph{Generalization limits} of the data. Despite the low results in Figure 6a we can see an increasing tendency over the years where the limits informed in 2022 (12,35\%) are higher than in 2016 (3,3\%). Looking closer at those data papers that disclose generalization limits, we observed that the informed limits relate to different aspects of the dataset. For instance, some of them point to specific non-recommended uses such as a climate dataset that states: \emph{"data presented here are not suitable for trend analysis --(at the US-Mexico and Canadian borders)--, since they use many stations that do not span the full temporal period 1950–2013}"\cite{livneh2015spatially}. On the other hand, some data papers point to specific procedures to use the data properly. For instance, a dataset for climate forecasting in East Africa states: "\emph{Users should take into account CenTrends' spatial error information when evaluating inter-seasonal and multi-annual rainfall anomalies. This type of information is poorly conveyed by cross-validation}" \cite{funk2015centennial}. Also, we found data papers discussing the limits of the theoretical approach of the data, such as a dataset about human footprint in environments stating: "\emph{Our work is subject to... limitations. First, like all attempts to map cumulative pressures, we did not fully account for all human pressures...}"\cite{venter2016sixteen}.

We also found disclosed limits linked to the gathering processes. For instance, a dataset  to develop and test movement recognition points to specific issues during the collection process: "\emph{In any use involving the amputated subjects (database 3), the users should keep in mind that in two amputated subjects (subjects 7 and 8) the number of electrodes was reduced to ten due to insufficient space and that three amputated subjects (subjects 1, 3 and 10) asked to interrupt the experiment before its end due to fatigue or pain}"\cite{atzori2014electromyography}. Finally, there are also limitations in terms of the annotation process. For instance, a genome dataset points at the use of an automatic algorithm to annotate the data: "\emph{the TOBG genomes have been generated using an automated process without manual assessment, and therefore, all downstream research should independently assess the accuracy of genes, contigs, and phylogenetic assignments for organisms of interest}" \cite{tully2018reconstruction}.

% Social Concerns
\textbf{Social concerns:} Out of the data papers representing people (16,5\% of the total sample), only 12,35\% are disclosing \emph{Social Concerns} of the data. %In that sense, in Figure \ref{fig:social}, we see an increasing tendency from 2016. The results are equivalent to the generalization limits, 
Curiously enough, datasets involving people do not tend to be better documented than the average, at least in terms of data limitations. Looking closer at the data papers disclosing social concerns, we observed that mentions to social biases are the most common, followed by privacy and sensitivity issues. As an example of a social bias, the Columbia Open Health Data states: "\emph{clinical databases contain a base population biased towards people with higher levels of existing conditions, thus biasing the measurements to overestimate the prevalence relative to the general population}"\cite{ta2018columbia}. 

\textbf{Maintenance policies and ML tested dimensions:} Data paper almost never discuss recommended \emph{Maintenance policies}; we only detected a 0,42\% (17). Looking at the few data papers that do cover this aspect, policies are basically restricted to the the updating timeframe for the data \cite{brlik2021long} and the maintenance of the sensor and instruments gathering the data \cite{tang2023chinese, kohler2022meteorological}. In terms of papers documenting ML models that have already been tested on the data, only the 8,81\% report this information. However, looking at Figure 6c we can see a clear growing tendency  %with being 2022 (15,1\%) clearly much higher than 2015 (5,1\%), 
probably due to the popularization of ML technologies.

% Collection
\textbf{Profile of the collection team and the collection targets:} In terms of collection processes, the \emph{Profiles of the collection team} is present only in 3,91\% of the data papers, and looking at Figure 6d, we can observe that the tendency over the years tends to be flat. Beyond the low presence, if we look closer at those data papers profiling the team, they often do it in the author's information section, disclosing only basic information such as affiliation, provenance, and job level. In addition, we also detected a subset of them being false positives as, instead of profiling the collection team, they were actually profiling the  target of the collection process, much more common (64,15\%) in data papers that are about people. %, and looking at Figure \ref{fig:coltarget}, we can see that there are no changes over the years. 
The high presence of this dimension could be explained as profiling targets' demographics is a mature practice in the healthcare field (e.g., the demographics of patient cohorts are crucial to include them or not in experiments or treatments). However, we also found complete profiles of the targets in data from other field such as analyzing factors influencing trading on pedestrian \cite{ajakaiye2018datasets}, or exploring the time perspective of Swedish citizens \cite{garcia2016time}.

\textbf{Speech context in language datasets}: This dimension was only present in 17,82\% of the relevant subsample though with a slight uptake in recent years (Figure 6f).  The need for documenting the speech context was firmly stated by Bender et al. \cite{bender-friedman-2018-data} in 2018, being one of the potential causes of this change in the tendency. Some examples of datasets disclosing proper speech context are Arabic diversified audio datasets \cite{lataifeh2020ar}, or a speech corpus for Quechua Collao \cite{paccotacya2022speech}, however, the information is mixed along the data papers and no structured form is provided as Bender at al.\cite{bender-friedman-2018-data} initially states. 

\textbf{Profile of the annotation team and annotation infrastructure and validation}: Finally, regarding the annotation process, we can see that the \emph{Profiles of the annotation team} is present only in 3,56\%  of the data papers (more informed than the collection team), but looking at Figure 6g, an contrasting to the collection process, it is improving (1,5\% in 2015 to 9,8\% in 2023). On the other hand, the \emph{Annotation Infrastructure} is present in 73,66\% of the sample, and information about \emph{Annotation Validation} methods is present only in 45,80\% in the analyzed sample. In both cases, the tendency across years is flat, without any relevant changes.

\textbf{Variation across topics.} Looking at the presence of the dimension across the different topics presented in the last section, we see that the reporting of specific dimensions varies depending on the topic. For instance, the dimensions regarding the profiling of the teams behind the annotation and the data collection are more commonly reported in Medical imaging data papers. In this topic, 8,64\% of the data papers report the annotator's profile versus 3,56\% of the entire sample, and 8,79\% report the collector's profile versus 3,91\% of the entire sample. On the other side, regarding data paper that reports tests with ML systems, we have seen that Human movement recognition, 30,11\%, and  Medical imaging, 19,22\% data papers are those that most commonly report them, closely followed by Agriculture \& Cropland, 16,15\% and Climate and hydrological analysis, 15,81\% data papers. In contrast, more commonly reported dimensions, such as the sources and the infrastructure or the description of the collection process, remain equivalent across topics. The data accompanying this study share the complete results and analysis regarding the dimensions' presence across topics.

\subsection{Dimension's presence across journals}

\begin{figure}[b!]
    \centering
    \includegraphics[width=1\textwidth]{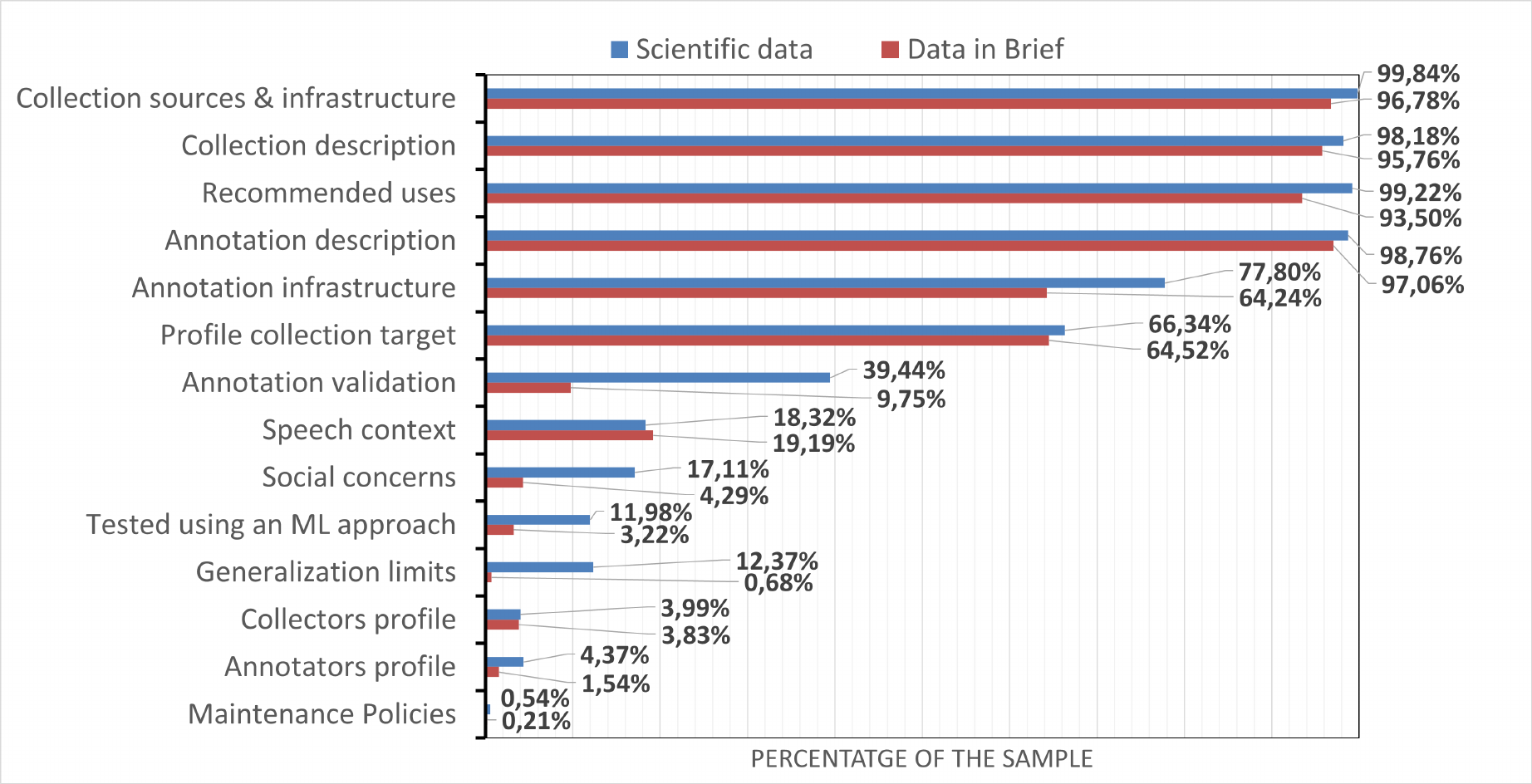}
    \caption{\label{fig:papercompare}  Overall results of informed dimensions for Data in Brief and Scientific Data. }
\end{figure}
In this section, we compare the two analyzed venues, Data in Brief and Scientific Data, comparing the dimension's presence in each of them. Figure \ref{fig:papercompare} shows an overview of the dimension's presence for each venue. In that sense, we have seen clear differences in the documentation practices of both journals. While the most common documented dimensions remain similar between venues, the less documented dimensions show clear differences between them. For instance, \emph{Generalization limits} is present in 12,37\% of data papers in Scientific Data, while is practically non-informed (0,68\%) in Data in Brief. A similar tendency is shown in the \emph{Social Concerns} dimensions where Scientific Data is informed 17,11\% while Data in Brief is just 4,29\%. On the other hand, looking at Figure \ref{fig:papercompare}, we can see a clear difference between the number of papers \emph{tested using ML approaches} between both venues; while in Scientific Data is 11,98\%, Data in Brief is only 3,2\%. However, looking at this dimension between years, we see a clear increasing tendency from both in the last years.

 \begin{figure}[b!]
    \centering
    \includegraphics[width=0.85\textwidth]{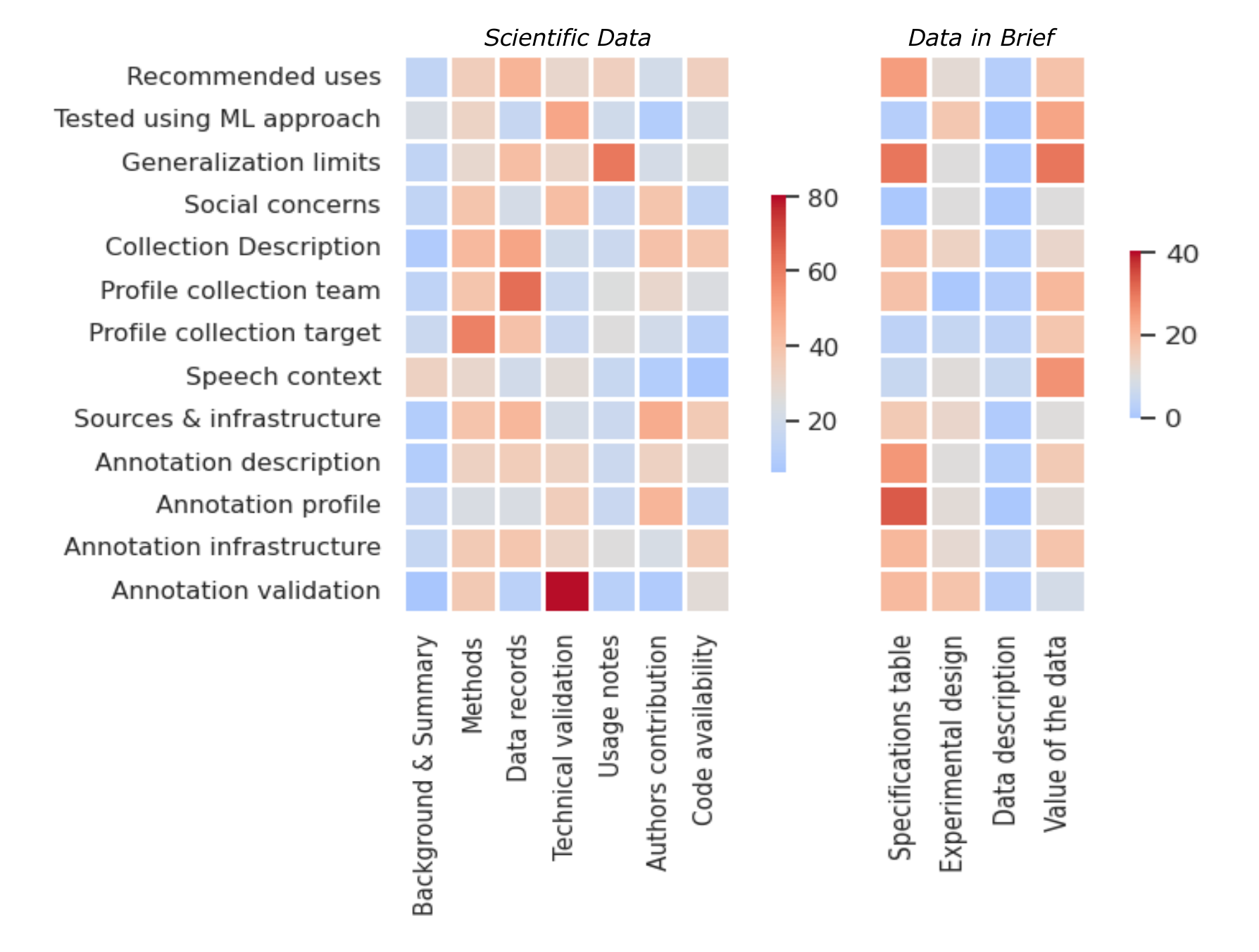}
    \caption{\label{fig:papersections}  Role of the recommended sections of the analyzed journals in relation with the dimension's presence. For each dimension (horizontal), each cell shows the percentage of manuscripts that contain relevant paragraphs retrieved from each recommended section. }
\end{figure}

Regarding the collection and annotation of the data, we see that the dimensions of the collection process remain similar between both venues. Specifically, the dimensions related to the data collection, such as the profile of the collection team, the collection target, or the speech context, show a similar presence between both venues. In contrast, the dimensions related to the annotation process show notable differences.  While the description of the process and the infrastructure share a similar presence, the validation methods show a clear difference (39\% and 9,75\%), and the profile annotation team seems to be slightly more present in Scientific Data than Data in Brief. 

To gain deep insight into the similarities and differences between the analyzed journals, we examined the role of the venue's recommended sections \cite{li2022data} for the dimensions under study. To do so,  we analyzed the results of the retriever model for those data papers that were informed with each specific dimension, looking at which sections corresponded to the most relevant retrieved paragraphs. Figure \ref{fig:papersections} shows which venue's sections are more likely to contain relevant information about each dimension. For instance, in Scientific Data's manuscripts where the \emph{Generalization limits} dimension was informed, 61,12\% of the manuscripts have relevant paragraphs retrieved from the \emph{Usage notes} section. 
The first thing to note is that the recommended sections differ between the venues, and the recommended structure's follow-up level is higher in Scientific Data than in Data in Brief, however the follow-up level in Data in Brief is enough to get insights about the sections' usage. 

Looking at the role of the sections across dimensions (vertically in Figure \ref{fig:papersections}), we can see that in Scientific Data, the \emph{Methods} and \emph{Data records} sections usually contain relevant information about all the analyzed dimensions. Besides, the \emph{Technical validation} and \emph{Usage notes} sections tend to contain more relevant information about the data's uses, limits, ML tests, and social concerns.
Similarly, the Data in Brief's \emph{Specification Table} and \emph{Value of the Data} sections are common across all dimensions. 
On the other hand, sections such as \emph{Background \& Summary} or \emph{Data description} are less likely to contain relevant information about the demanded dimensions. Looking at each specific dimension (horizontally in Figure \ref{fig:papersections}), we can see that some reported dimensions are very frequently reported in particular sections. For example, in the case of \emph{Generalization limits}, the Scientific Data's \emph{Usage Notes} and the Data in Brief's \emph{Specification Table} and \emph{Value of data} sections typically contain more relevant information. The \emph{Annotation profile} is typically reported in the Data in Brief's \emph{Specification Table} section. Finally, the annotation validation methods reported in Scientific Datra are always practical in the \emph{Technical Validation} sections.

\subsection{Comparing the dimension's presence with ML-focused venues}
\label{neurips}
\begin{figure}[b!]
    \centering
    \includegraphics[width=0.95\textwidth]{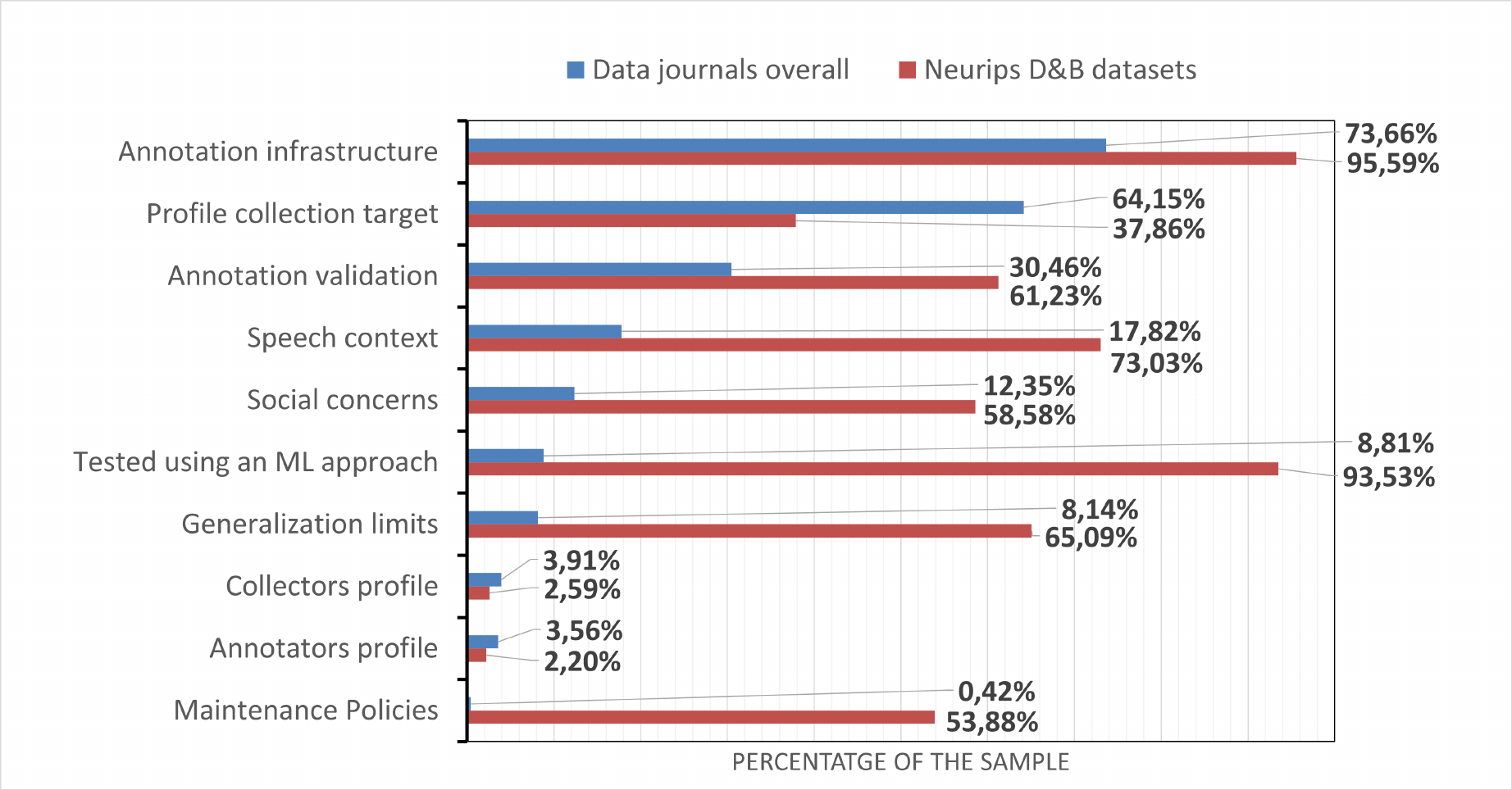}
    \caption{\label{fig:paperneurips}  Comparison between the obtained results and the dataset published in Neurips Dataset \& Benchmark track (2021-2023)}
\end{figure}

To compare the analyzed sample with an ML-focused venue, we analyzed the publication describing datasets of the NeurIps Datasets and Benchmark track, corresponding to 232 data papers (paper presenting a dataset exclusively from 2021 (year of track’s creation) to 2023 (2024 still not published). The Dataset and Benchmark track is the answer for the ML community to build data-centric ML practices. It is intended to publish datasets and data-practice papers focused on improving the state-of-the-art of ML models. The track encourages authors to present a manuscript similar to that of data journals and an attachment documenting the dataset. An interesting point is that the NeurIps D\&B encourages authors to include dataset documentation following the frameworks presented in the background, which has served as a basis for our work. 90.95\% of the analyzed publication contains an attachment corresponding to this documentation assets, and these attachments has been merged into the original manuscript for its analysis. In Figure \ref{fig:paperneurips}, we show the presence of the least documented dimension in our sample together with the results we got in our analysis.

The results show notable differences between the dataset published in the NeurIps Datasets and Benchmark track and the data journals. While in data journals, generalization limits stay at 8.14\% and social concerns stay at 12.35\%, in NeurIps Datasets and Benchmark track, we see that generalization limits rise to 65.08\%, and social concerns rise to 58.36\%. This difference is even more significant in the case of maintenance policies, where in data journals, this dimension is practically missing (0.42\%), while in Neurips, this rises to 53.88\%. Similar things happen with language datasets documenting the speech context: In journals, it is 17.21\%, and in NeurIps Datasets and Benchmark track, it is 72.22\%. Finally, and as an evident outcome, the number of publications tested with an ML technique is very high (93.53\%).\looseness=-2

In addition, we can observe that the annotation process solidifies and is much more documented in the ml-focused track. In that sense, the documentation of the infrastructure goes from 73.66\% in the data journals to 95.59\% in the Neurips Datasets and Benchmark track. On the other hand, the annotation validation methods also experience an apparent rise, going from 30.46\% in the journals 
to 61.23\% in the Neurips publication. In general terms, we have observed that the annotation process has more relevance in ML-focused venues, where practically all the data papers (98.28\% in contrast to the 42.28\%) present some sort of annotation process.

However, the dimensions regarding the profile of those involved in the creation of the dataset are similar and even less present than in scientific data journals. In that sense, the scarce presence of the profile of the collection team (2.59\% vs. 3.91\%) or the annotation team (3.56\% vs. 2.20\%) shows that these dimensions are still a challenge, and the recommended documentation frameworks are not enough to encourage the community to document it. On the other hand, the profile of target people of the collection process (64.15\% vs. 37.6\%) is less present in ML-focused venues, probably due to this practice being significantly extended in healthcare datasets, that represent a small proportion of manuscripts in the analyzed venue.

\section{Discussion}
\label{sec:dis}

This section discusses additional conclusions from our analysis and presents a set of recommendations to increase the presence of the demanded dimension in scientific data documentation. We also discuss current metadata infrastructure challenges and future trends in scientific data discoverability.

\subsection{Improving submission guidelines}

One of the first insights from the results presented in the last section is the contrast between the most and least documented dimensions. We have observed that the most documented dimensions are explicitly mentioned in the authors' submission guidelines of the data journals. For instance, Scientific Data points the following regarding \emph{Data uses}; \emph{"[...] briefly outline the broader goals that motivated the creation of this dataset and the potential reuse value... \cite{sdata}"}. As another example, in disclosing information about data sources, Data-in-Brief states \emph{"Please mention where the data were collected (e.g., geographical coordinates) or where the data are stored (typically your affiliation)... \cite{dbrief}"}. In conclusion, we observed that the \textbf{publisher's guidelines are a strong tool to ensure the documentation quality of datasets} as the research community follows them strongly. 

Another clear indication that submission guidelines are beneficial comes from a comparison with the NeurIps Dataset and Benchmark track, where the recommendation to follow the documentation frameworks supplied in the background has significantly increased some of the dimension's presence. However, we observed that most of the published documentation is sparse and follows distinct frameworks; even when following the same frameworks, the dimensions reported vary.  In that regard, given the distinct nature of data journals in comparison to an ML-focused venue, as well as the fact that they target a broader spectrum of scientific fields, we see improving the already existing and mature submission rules as the most effective approach of enhancing data documentation. 

Therefore, we present a set of specific suggestions for data journals, some based on the frameworks, to improve the current guidelines aimed at strengthening the presence and completeness of the less documented dimensions.

\subsubsection*{Guidelines for \emph{data generalization limits} and \emph{social concerns} }

As we observed in the last section, the \emph{data generalization limits} affect all stages of the data creation process. %In that sense, we observed limits forbidding specific uses of the data or providing strong recommendations on how to use the data properly. But, limitations also come from data provenance aspects. For instance, we observe limitations regarding the sources (pointing to the lack of quality or data distribution of these sources) and coming directly from the gathering and analysis processes. On the other hand, we observed that \emph{Social concerns} and \emph{Data generalization limits} are very interrelated as a generalization limit of a dataset representing people could be easily expressed as a social concerns.
We propose asking authors about the generalization limits of their data using a structured report highlighting the various aspects we observed in this study. Furthermore, in the case of datasets gathered or representing people, we propose adding specific aspects pointing to social group biases as well as sensitivity and privacy issues. Table \ref{recommendations} summarizes the recommendations for the dimensions of data generalization limits and social concerns. In addition, beyond the limitations aspects identified in this study, \textbf{a consistent taxonomy of the data limitation aspects would be a useful tool for data creators to reason about the generalization limits of their data}. In that sense, building this taxonomy seems a promising research line with a potential impact on improving data-sharing practices.

\subsubsection*{Guidelines for  \emph{profiling the teams} behind the data creation process }

Recent works in the ML community state that profiling the teams involved in the dataset creation in its documentation is essential. The main reason to do that is transparency, but also, having information about who is behind the data creation gives users more confidence and trust in the system \cite{chen2023ai}. Also, as a particular case of the team's type, the labor work behind the data has raised concerns in the ML community\cite{barbosa2019rehumanized, diaz2022crowdworksheets} about the quality of the data and the worker's labor conditions. Despite this, we observed that the teams involved in the different provenance processes are scarcely documented, also in the ML-focused venue, and there is a need to find solutions to this situation. In that sense, to engage authors to profile the different teams participating in the dataset creation, we propose to provide a generic \emph{Data participant profile}, shown in Table \ref{recommendations}, for every provenance process (annotation, collection, preprocessing) that involved people, along with the submission guidelines.

\subsubsection*{Guidelines for annotation processes}

We observed that annotation processes are worse informed than gathering processes. For instance, at least 26,34\% annotated datasets do not mention the infrastructure they use to annotate them, where gathering processes are practically always informed in that sense. The same occurs with the label validation methods, where practically 41\% of data papers with an annotation process do not mention any. To improve this situation, we propose to add specific statements, shown in Table \ref{recommendations} inside the methods section of the submissions guidelines. With these statements, we aim to \textbf{encourage authors to disclose the infrastructure and the validation methods applied over the labels, as well as the annotations guidelines given to the annotator team}.

\subsection{Explicitly covering new dimensions for the future use of scientific data, especially in ML settings}

To better prepare for the long-term use of the published datasets we suggest two specific improvements in the data papers documentation guidelines. 

\subsubsection*{\emph{Maintenance policies} for evolving the data}

Data has an evolving nature, as the reality it aims to represent. Despite this, we found a few papers outlining data maintenance policies. Data papers rarely contain information about how often the data will be updated, how users can contribute and report errors found in it, and how authors will fix them. The most common informed updating policies are an informal commitment of the authors, such as Mahood et al. \cite{mahood2022country} "\emph{Will be actively maintained and updated yearly and upon request... by the authors}". This situation contrast with the results we obtained when analyzing the NeurIps D\&B publications, when nearly 53,88\& contains information about the maintenance policies. Also, it contrasts with  the recent ML community works, such as Luccioni et al.\cite{sasha} and Gebru et al.\cite{gebru2021datasheets}, which states the importance of documenting these practices. In these works, the authors stated that datasets could be updated or even deprecated for many reasons, such as legal enforcement or the apparition of ethical issues in the data (such as the case of ImageNet \cite{10.1145/3351095.3375709}).

In our view, one of the main reasons this situation happens is the inherent static form of scientific publication. \textbf{The data papers' static form may be limited to represent the nature of the dataset}. However, other types of scientific publications, such as the \emph{Original Software Publication}, have adopted an evolving nature, encouraging authors to update and evolve their published artifacts. We propose a similar approach for scientific data papers where authors can state the current status of the data artifact and guidelines for their future evolution.  

In Table \ref{recommendations} we outlined the main concepts of the \emph{Maintenance policies} dimensions.

\subsubsection*{Adding data licenses and structured ML experiments results}

Even if the number of data papers, as shown in Figure 6c, that are designed with ML in mind is still low, we can be sure that most (if not all) of those datasets will be part of some ML pipeline in the future due to the data-hungry tendency of ML. So, we need methods to let authors think about the potential use of their data in ML approaches and methods to protect their rights.

% Reflexion

In this sense, we propose \textbf{the integration of the Montreal Data License \cite{benjamin2019towards} template in the submission form}. This license template helps authors to decide how to disclose the data's rights towards its use in ML technologies in a structured way. It also helps to disclose the authors intentions in the use of the data in ML. Data could not be suited for ML for many reasons, and potential ML practitioners need to be aware of authors' thoughts on this. We would also like to encourage authors to test themselves their data using ML approaches. For this,  we propose to \textbf{annotate the results of ML experiments in a structured form} as shown in Table \ref{recommendations}. A structured form can be later included in the publishers metadata, enhancing the discoverability of the data through search engines (as Google Dataset Search via Croissant \cite{akhtar2024croissant} format), and facilitating the inclusion of the data in popular ML platforms such as HuggingFace or Kaggle.

\setlength{\extrarowheight}{1.6pt}
\begin{table}[t!]
\small
    \centering
    \caption{Overview of the recommendation guidelines for each dimension }
    \begin{tabular}{p{4.5cm} p{12cm}}
        \multicolumn{2}{l}{\textbf{Data generalization limits and social concerns}}   \\ \hline
         Recommendations: & How to use the data properly and specific non-recommended uses. \\
         Design limitations: & Data limits regarding data distribution and composition, and the theoretical approaches behind data design. \\
         Provenance \& source limitations: & Limits generated by the collection/annotation or inherited from the data sources \\
         \\
         \multicolumn{2}{l}{\textbf{Maintenance policies} (To be released in every version)} \\ \hline
         Maintainers \& updating timeframes: & Who are the current maintainers of the dataset and how can be contacted. Is there a plan to update the dataset with new data or with corrections. \\
        Erratum \& contribution guidelines: & How errors can be reported, and how people can contribute to the dataset\\
        Deprecation policies: & Which are the current plans for deprecating the dataset\\ \\
        
         \multicolumn{2}{l}{\textbf{ML approaches using tested with the dataset} } \\ \hline
         Machine learning task: &  Specific task description (character recognition, pose estimation, classification, etc.).\\
         Machine-learning model \& metrics: & The ML model name trained with the data and metrics (F1, accuracy, precision, recall).\\
         References: & The reference to source code and relevant leadebords.\\ \\
         
         \multicolumn{2}{l}{ \textbf{Data participants profile} (For each provenance process involving people)} \\ \hline  
                Participants: & Number of the team members and high-level description of the team.\\
                Demographics information: & Age, gender, country/region, race/ethnicity, native language, socioeconomic status, etc.\\
                Team type and labour information: & Internal, external, crowd-sourcing or  citizen science (volunteers) teams. If team is crowd-sourcing authors should disclose labour information, such as range salary, country of the contractor, labour union policies, etc. \\
                Qualifications: & Members training and qualifications in relevant disciplines regarding the specific task applied over the dataset.\\
        \\ 
        \multicolumn{2}{l}{ \textbf{Annotation guidelines, infrastructure and validation}} \\ \hline  
         Annotation guidelines: &  The guidelines shared with the annotator team and how the annotation can be reproduced.\\
         Infrastructure: & Which infrastructure (software, or physical) have been used to label the data.\\
         Validation methods: & Validation methods applied over the labels (e.g., inter-annotation agreements).\\
         
    \end{tabular}
   
    \label{recommendations}
\end{table}

\subsection{Metadata is not enough, text analysis tools are still needed}

\begin{figure}[b]
    \centering
    \includegraphics[width=8cm]{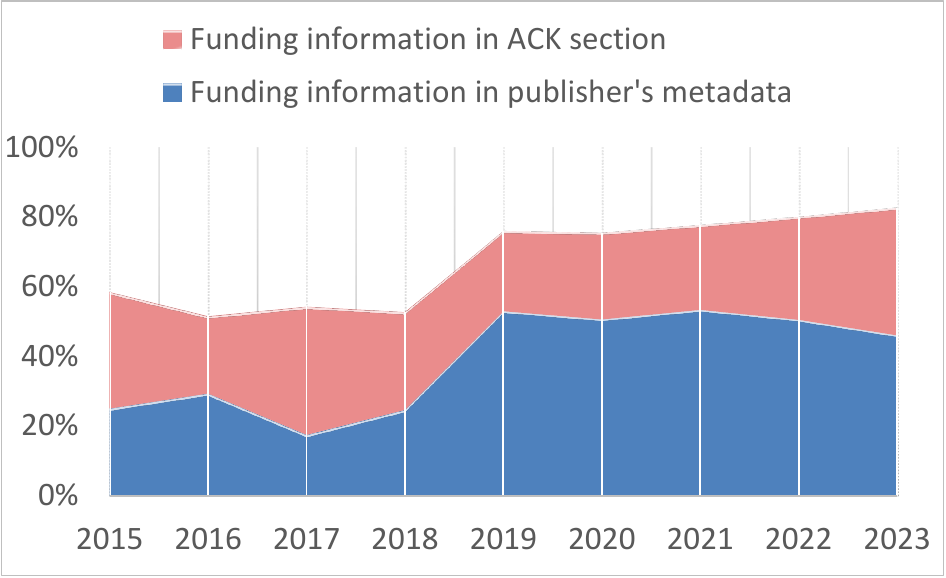}
    \caption{\label{fig:funders} Funding information in publishers metadata, and extracted using text analysis over the paper's acknowledge section.}
    \end{figure}

Unfortunately, metadata may be wrong or incomplete, even for the most common fields. For instance, one of the dimensions typically found in metadata is the funding information. Funding agencies often request this information as proof of the research to justify the grants, and authors are motivated to disclose this information within the submission process. But if we compare the percentage of papers disclosing funding information (Figure \ref{fig:funders}) we see that this number is smaller than the number we can detect by running a text analysis on the Acknowledgments section. Indeed, although metadata for annotating funders has been implemented over the years, and it is supposed to be mandatory to justify research funding, we found that this does not consistently match the ones in the text manuscript. Recent works \cite{zhang2024missing} have pointed to metadata quality issues with scholarly catalogs (as OpenAlex) as one potential cause of these inconsistencies. Moreover, these issues show the need to complement all the existing and future initiatives to put in place better guidelines for authors with text analysis methods such as the one we used in this same paper to automatically enrich the paper metadata.

\subsection{Machine-readable description of the dimensions to improve data discoverability}

Searching and evaluating the suitability of a dataset for a particular use case is a difficult task. As stated in Section \ref{sec:back}, many aspects could influence the quality of the ML models trained with it, such as the demographics of the collection and annotation team or the particular generalization limits of the data. In that sense, 
search queries for datasets with annotators of the same geographic zone or gathered from a specific kind of patient are still complex to resolve using the current data search engines and data repositories.

In response to this situation, recent initiatives such as DescribeML \cite{ginerDSL} and Croissant \cite{akhtar2024croissant} are proposing structured ways for documenting datasets for machine learning. These structured formats are built on top of the works we presented in Section \ref{sec:back} and facilitate the automatic search and analysis of the presented dimensions. Moreover, popular data search engines, such as the Google Dataset Search \cite{brickley2019google}, are starting to integrate these initiatives, showing their potential in the future of the data discoverability. Therefore, and beyond improving the publisher's submission guidelines, we propose capturing the recommended dimensions in a structured format (compliant with some of the standards and initiatives mentioned before), for instance, during the data paper's submission process. \looseness=-1

\section{Methods}

\subsection*{Sample Selection}

The efforts for improving data-sharing practices in the scientific field has trigger the creation of research data journals and tracks, e.g. see these compilations of data journals \cite{jiao2023exclusively, candela2015data, walters2020data}, devoted to publish mainly data papers (more than 50\% of their total publication volume). To build our data sample, we selected from this list the journals meeting the following criteria: 1) Journal is active at July of 2023 and publish data papers regularly, 2) it publishes data papers from different scientific fields (inter-disciplinary scope), 3) it publishes data papers written in English. Filtering by this criteria, we have selected \emph{Scientific Data} and \emph{Data in Brief} as candidates. Scientific Data published by Nature was funded in 2014, with 3100 data descriptors published so far and covering the whole range of natural sciences disciplines. Data-in-Brief is published by Elsevier, also funded in 2014 and with 9315 data notes published from all the scientific disciplines.

\subsection*{Collecting the sample of data papers}
 
For this study we analyzed the full manuscripts of the data papers. To collect such manuscripts we relied on OpenAlex API \cite{priem2022openalex}, a major database commonly used in large-scale bibliographic analysis, which, recent studies (Jiao et al\cite{jiao2023exclusively}), has pointed as one of the most complete databases in indexing data papers. To separate the data papers from other regular papers, and given that the own paper classification was not reliable (data papers are often indexed as regular papers), we develop an in-house python script (also available in our repository) to check one by one the type of the paper in the publisher's website. The specific page for each paper was derived from its DOI. From the obtained list of data papers (3100 in Scientific Data and 9315 in Data-in-Brief) we selected a random sample, of data papers published between 2015 and July 2023, ending up with a sample of 4041 data papers, (2549 from Scientific Data and 1492 from Data-in-brief). Finally, we got the full manuscripts in PDF format using the DOI and accessing again the publisher’s website. 

In terms of the publication of Neurips Dataset and Benchmark track, we got all the publications from the venue from the OpenReview API together with the supplementary material attaches. An in-house script has filtered those that do not present a data papers, and has attached the potential dataset documentation (if existing) as an annex to the main manuscript. From this steps both kind of manuscripts, those from data journals and form Neurips have followed the same steps.

\subsection*{Manuscripts and text preparation}

To prepare the texts for the automatic extraction of dimensions data, we have parsed each manuscript  using the SciPDF library \cite{scipdf}, that employs the GROBID\cite{romary:hal-01673305} service under the hood. From the parsed documents we excluded the headers, footers, and the references section. Then we divided each section in different chunks composed of the title and text of the sections with a maximum of 1000 words (600-700 tokens). For longer sections, we created a new chunk of the section with an overlap of 300 characters with the previous chunk and also starting with the section title to provide context to the chunks. We have implemented this chunking strategy because we observed that the rhetoric structure in scientific articles provides good context and improves the performance of language models in retrieving relevant passages regarding to specific questions.

For the figures and tables present in each data paper we used Tabula.py \cite{tabula} to extract the data together with its caption. Figures caption has been added as another chunk to the ones created previously. Then, we parsed the tables to HTML, and along with the caption and the paragraphs referencing the table, we used a Language Model (text-davinci-003) from the OpenAI API service to generate an explanation of the tables content using the table. This explanation then was added as yet another chunk. It has been useful to detect if some dimensions, such as the demographics of the team, where maybe present in the tables content.

\subsection*{Dimensions extraction and analysis}

\begin{figure}[b!] 
\centering
 \includegraphics[width=0.8\columnwidth]{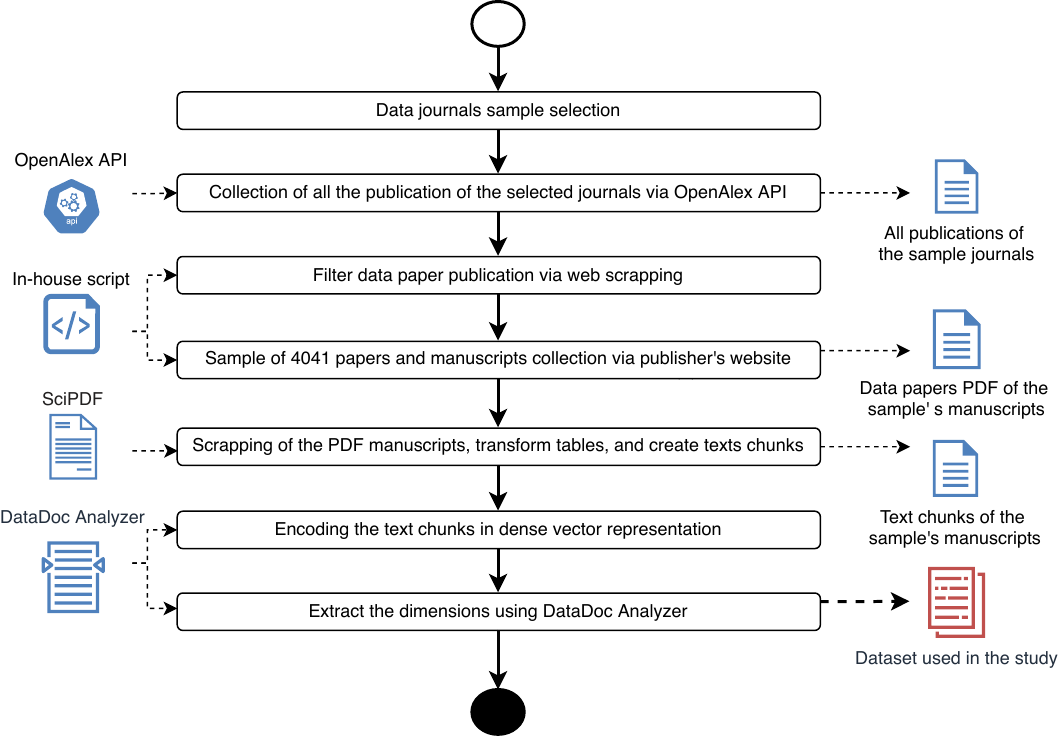}
 \caption{Data collection and preparation workflow}
 \label{fig:ui}
\end{figure}

To extract the dimensions presented in Section \ref{sec:back} we applied the method presented in \cite{datadoc}, which implements a Retrieval Augmented Generation (RAG) approach specifically targeting these dimensions. It works by combining a retrieval model and a chain of prompts on top of the retrieved paragraphs. Both, the identification of the candidate paragraphs and the type and concrete verbalization of the prompts, are tailored to each individual dimension to maximize the chances of detecting it. As a final step, a zero-shot classification strategy, also presented in the previous work, using a fine-tuned version of \mbox{BART \cite{lewis2020bart}} on the MultiNLI (MNLI) dataset \cite{williams2018broad} is used to determine whether the result of the chain includes the target dimension. For assessing the type of team and the type of annotation or collection process, the method performs a last step in the chain's prompt, asking to classify the obtained answer within a set of predefined values. The predefined types of annotation and collection processes were heuristically designed by the authors, looking at a subsample, and is intended to evaluate the diversity of the sample in terms of type of process. This whole process has been applied to each preprocessed manuscript to generate the data points used in our analysis. Figure \ref{fig:ui} summarizes these steps.

In addition, we made a topic analysis of the generated uses explanations using BERTopic \cite{grootendorst2022bertopic}. In that sense, we did a semi-supervised analysis, generating the topics from the data, and then reducing it and cleaning non-relevant ones. The representation of the topics shown in Table 1 has been generated with a fine-tune layer using a language models (flan-t5-large). The full results are published together with the data.

In conclusion, the data used to perform this study is composed of a list of the 4041 data papers, enriched with the dimension extracted and with the results of the topic analysis. As Giner et al.\cite{datadoc} reports in their approach, language models have a tendency to hallucinate, and despite the different strategies implemented to reduce this issue, we need to keep this in mind when looking the data. In that sense, the report of his study have reported preliminary accuracy metrics for each dimensions, being  of 88,26\% for the uses, 70\% of the collection, and 81,25\% for the annotation dimension.

\subsection*{Extraction validation}

The extraction approach is powered by large language models (GPT3.5) to extract the demanded dimensions. Despite the different strategies implemented to reduce this issue, we should point out that these models tend to hallucinate when looking at the data. In this regard, the authors of the method give preliminary accuracy metrics for the demanded dimension, assessing the quality of the answers: 88.26\% for the uses, 70\% for the collection, and 81.25\% for the annotation dimension \cite{giner2024using}. Furthermore, the authors stated that regardless of the quality of the answer, the models performed well in recognizing whether or not the information was contained in the paper, being this the specific task used in our study. 

To thoroughly understand the method's accuracy in the specific task used in this study, we performed a manual evaluation of a subsample of our data (about 1\%). In that sense, we annotated whether the demanded dimensions were present or not in the data paper, and then we compared the results of the extraction with the manual annotations. The sample have been random extracted from the data and the results shows comparable distribution of dimension presence to the overall data.

The validation results demonstrate good performance in detecting the presence of the required dimensions. For example, data limits and social concerns (94.59\%), data tested using an ML technique (97.30\%), and the demographic profile of the collection team (97.10\%) and target (83.33\%) were practically always detected. However, the method faced more difficulty recognizing whether the paper had an annotation process (83.78\% accuracy). Some errors were caused by misinterpreting aspects of the collecting process as annotations (e.g, “\textit{biological agents were counted using a Sedgewick Rafter cell and a microscope}”) or data management operations as annotations (e.g, “\textit{all recordings were segmented and labeled with an id}”). Also, in line with Giner et al. \cite{giner2024using}, the identification of validation techniques had the lowest accuracy (72.22\%) because label validations were often confused with validation methods during data collection. In resume, while the method's accuracy is far from perfect, we consider it enough for the analysis presented in this paper. The data and the complete analysis of the detected errors during the validation are published together with the data.

\section*{Data availability}

The data used in this study is published in an online repository and archived in Zenodo with a permanent identifier to facilitate its findability \cite{zenodoRepo}. The data maintainer is Giner-Miguelez, co-author of this work, and can be contacted for reporting issues (errata) or further contributions to the data. Issues and contributions can also be reported through Zenodo's open repository link. There is no set timeframe for data updates, and any data deprecation will be announced in the Zenodo-linked repository. As data does not represent people directly, we cannot infer potential social concerns directly from it. Aside from the topic analysis, we did not test the data with any machine learning approach. However, because data can be used to train ML models, practitioners should be aware of the limitations presented before using it.

The data extracted using the Giner et al. \cite{datadoc} approach cannot be understood as a gold truth because the underlying models are limited in accuracy and may suffer from hallucinations. Because of these issues, we only recommend using the data for statistical analysis of its trends. The data sources were the OpenAlex API \cite{priem2022openalex}, the OpenReview API service,  and the data publishers' website (Nature's Scientific Data and Elsevier's Data in Brief), and we have not detected any explicit bias inherited from them. The data collection and curation were done by the three authors of the paper (internal team), who are based in Spain and Luxembourg, during 2023. The method section explains the provenance processes carried out by the authors, and the archived repository contains the calculus and charts generated from the data.

\section*{Code availability}

The code used to perform the analysis of this work has been published in the same public repository as the data along with the instructions to reproduce the extraction process of the data.

\section*{Acknowledgements}

This research has been partially supported by the Spanish government (LOCOSS - PID2020-114615RB-I00), the AIDOaRt project, which has received funding from the ECSEL Joint Undertaking (JU) under grant agreement No 101007350. The JU receives support from the European Union’s Horizon 2020 research and innovation programme and Sweden, Austria, Czech Republic, Finland, France, Italy, and Spain. Jordi Cabot is supported by the Luxembourg National Research Fund (FNR) PEARL program, grant agreement 16544475. \looseness=-1

\section*{Authors contributions}

The authors of the manuscript have been contributed equally to this work.

\section*{Competing interests}

The authors declare no competing interests.

\bibliography{refs}

\begin{thebibliography}{10}
\urlstyle{rm}
\expandafter\ifx\csname url\endcsname\relax
  \def\url#1{\texttt{#1}}\fi
\expandafter\ifx\csname urlprefix\endcsname\relax\def\urlprefix{URL }\fi
\expandafter\ifx\csname doiprefix\endcsname\relax\def\doiprefix{DOI: }\fi
\providecommand{\bibinfo}[2]{#2}
\providecommand{\eprint}[2][]{\url{#2}}

\bibitem{genderimbalance}
\bibinfo{author}{Larrazabal, A.~J.}, \bibinfo{author}{Nieto, N.},
  \bibinfo{author}{Peterson, V.}, \bibinfo{author}{Milone, D.~H.} \&
  \bibinfo{author}{Ferrante, E.}
\newblock \bibinfo{journal}{\bibinfo{title}{Gender imbalance in medical imaging
  datasets produces biased classifiers for computer-aided diagnosis}}.
\newblock {\emph{\JournalTitle{Proceedings of the National Academy of
  Sciences}}} \textbf{\bibinfo{volume}{117}}, \bibinfo{pages}{12592--12594}
  (\bibinfo{year}{2020}).

\bibitem{abbasi2020risk}
\bibinfo{author}{Abbasi-Sureshjani, S.}, \bibinfo{author}{Raumanns, R.},
  \bibinfo{author}{Michels, B.~E.}, \bibinfo{author}{Schouten, G.} \&
  \bibinfo{author}{Cheplygina, V.}
\newblock \bibinfo{title}{Risk of training diagnostic algorithms on data with
  demographic bias}.
\newblock In \emph{\bibinfo{booktitle}{Interpretable and Annotation-Efficient
  Learning for Medical Image Computing: Third International Workshop, iMIMIC
  2020, Second International Workshop, MIL3ID 2020, and 5th International
  Workshop, LABELS 2020, Held in Conjunction with MICCAI 2020, Lima, Peru,
  October 4--8, 2020, Proceedings 3}}, \bibinfo{pages}{183--192}
  (\bibinfo{organization}{Springer}, \bibinfo{year}{2020}).

\bibitem{zech2018variable}
\bibinfo{author}{Zech, J.~R.} \emph{et~al.}
\newblock \bibinfo{journal}{\bibinfo{title}{Variable generalization performance
  of a deep learning model to detect pneumonia in chest radiographs: A
  cross-sectional study}}.
\newblock {\emph{\JournalTitle{PLOS Medicine}}} \textbf{\bibinfo{volume}{15}},
  \bibinfo{pages}{1--17} (\bibinfo{year}{2018}).

\bibitem{datasheets}
\bibinfo{author}{Gebru, T.} \emph{et~al.}
\newblock \bibinfo{journal}{\bibinfo{title}{Datasheets for datasets}}.
\newblock {\emph{\JournalTitle{Communications of the ACM}}}
  \textbf{\bibinfo{volume}{64}}, \bibinfo{pages}{86--92}
  (\bibinfo{year}{2021}).

\bibitem{ginerDSL}
\bibinfo{author}{Giner-Miguelez, J.}, \bibinfo{author}{Gómez, A.} \&
  \bibinfo{author}{Cabot, J.}
\newblock \bibinfo{journal}{\bibinfo{title}{A domain-specific language for
  describing machine learning datasets}}.
\newblock {\emph{\JournalTitle{Journal of Computer Languages}}}
  \bibinfo{pages}{101209} (\bibinfo{year}{2023}).

\bibitem{tedersoo2021data}
\bibinfo{author}{Tedersoo, L.} \emph{et~al.}
\newblock \bibinfo{journal}{\bibinfo{title}{Data sharing practices and data
  availability upon request differ across scientific disciplines}}.
\newblock {\emph{\JournalTitle{Scientific data}}} \textbf{\bibinfo{volume}{8}},
  \bibinfo{pages}{192} (\bibinfo{year}{2021}).

\bibitem{wilkinson2016fair}
\bibinfo{author}{Wilkinson, M.~D.} \emph{et~al.}
\newblock \bibinfo{journal}{\bibinfo{title}{The fair guiding principles for
  scientific data management and stewardship}}.
\newblock {\emph{\JournalTitle{Scientific data}}} \textbf{\bibinfo{volume}{3}},
  \bibinfo{pages}{1--9} (\bibinfo{year}{2016}).

\bibitem{rolando2015data}
\bibinfo{author}{Rolando, L.} \emph{et~al.}
\newblock \bibinfo{journal}{\bibinfo{title}{Data management plans as a research
  tool}}.
\newblock {\emph{\JournalTitle{Bulletin of the Association for Information
  Science and Technology}}} \textbf{\bibinfo{volume}{41}},
  \bibinfo{pages}{43--45} (\bibinfo{year}{2015}).

\bibitem{mayernik2015peer}
\bibinfo{author}{Mayernik, M.~S.}, \bibinfo{author}{Callaghan, S.},
  \bibinfo{author}{Leigh, R.}, \bibinfo{author}{Tedds, J.} \&
  \bibinfo{author}{Worley, S.}
\newblock \bibinfo{journal}{\bibinfo{title}{Peer review of datasets: When, why,
  and how}}.
\newblock {\emph{\JournalTitle{Bulletin of the American Meteorological
  Society}}} \textbf{\bibinfo{volume}{96}}, \bibinfo{pages}{191--201}
  (\bibinfo{year}{2015}).

\bibitem{silvello2018theory}
\bibinfo{author}{Silvello, G.}
\newblock \bibinfo{journal}{\bibinfo{title}{Theory and practice of data
  citation}}.
\newblock {\emph{\JournalTitle{Journal of the Association for Information
  Science and Technology}}} \textbf{\bibinfo{volume}{69}},
  \bibinfo{pages}{6--20} (\bibinfo{year}{2018}).

\bibitem{kim2020analysis}
\bibinfo{author}{Kim, J.}
\newblock \bibinfo{journal}{\bibinfo{title}{An analysis of data paper templates
  and guidelines: types of contextual information described by data journals}}.
\newblock {\emph{\JournalTitle{Science Editing}}} \textbf{\bibinfo{volume}{7}},
  \bibinfo{pages}{16--23} (\bibinfo{year}{2020}).

\bibitem{faniel2019context}
\bibinfo{author}{Faniel, I.~M.}, \bibinfo{author}{Frank, R.~D.} \&
  \bibinfo{author}{Yakel, E.}
\newblock \bibinfo{journal}{\bibinfo{title}{Context from the data reuser’s
  point of view}}.
\newblock {\emph{\JournalTitle{Journal of Documentation}}}
  \textbf{\bibinfo{volume}{75}}, \bibinfo{pages}{1274--1297}
  (\bibinfo{year}{2019}).

\bibitem{zenodoRepo}
\bibinfo{author}{Giner-Miguelez, J.}, \bibinfo{author}{Gómez, A.} \&
  \bibinfo{author}{Cabot, J.}
\newblock \bibinfo{title}{Archived data supporting the study "\mbox{On} the
  readiness of scientific data papers for a fair and transparent use in machine
  learning"}, \url{10.5281/zenodo.10514145} (\bibinfo{year}{2024}).

\bibitem{mcmillan2021reusable}
\bibinfo{author}{McMillan-Major, A.} \emph{et~al.}
\newblock \bibinfo{title}{Reusable templates and guides for documenting
  datasets and models for natural language processing and generation: A case
  study of the {H}ugging{F}ace and {GEM} data and model cards}.
\newblock In \emph{\bibinfo{booktitle}{Proceedings of the 1st Workshop on
  Natural Language Generation, Evaluation, and Metrics}},
  \bibinfo{pages}{121--135} (\bibinfo{publisher}{ACM},
  \bibinfo{address}{Online}, \bibinfo{year}{2021}).

\bibitem{bender-friedman-2018-data}
\bibinfo{author}{Bender, E.~M.} \& \bibinfo{author}{Friedman, B.}
\newblock \bibinfo{journal}{\bibinfo{title}{Data statements for natural
  language processing: Toward mitigating system bias and enabling better
  science}}.
\newblock {\emph{\JournalTitle{Transactions of the Association for
  Computational Linguistics}}} \textbf{\bibinfo{volume}{6}},
  \bibinfo{pages}{587--604} (\bibinfo{year}{2018}).

\bibitem{holland2020dataset}
\bibinfo{author}{Holland, S.}, \bibinfo{author}{Hosny, A.},
  \bibinfo{author}{Newman, S.}, \bibinfo{author}{Joseph, J.} \&
  \bibinfo{author}{Chmielinski, K.}
\newblock \bibinfo{journal}{\bibinfo{title}{The dataset nutrition label}}.
\newblock {\emph{\JournalTitle{Data Protection and Privacy, Volume 12: Data
  Protection and Democracy}}} \textbf{\bibinfo{volume}{12}}, \bibinfo{pages}{1}
  (\bibinfo{year}{2020}).

\bibitem{sasha}
\bibinfo{author}{Luccioni, A.~S.} \emph{et~al.}
\newblock \bibinfo{title}{A framework for deprecating datasets: Standardizing
  documentation, identification, and communication}.
\newblock In \emph{\bibinfo{booktitle}{Proceedings of the 2022 ACM Conference
  on Fairness, Accountability, and Transparency}}, FAccT '22,
  \bibinfo{pages}{199–212} (\bibinfo{publisher}{Association for Computing
  Machinery}, \bibinfo{address}{New York, NY, USA}, \bibinfo{year}{2022}).

\bibitem{akhtar2024croissant}
\bibinfo{author}{Akhtar, M.} \emph{et~al.}
\newblock \bibinfo{title}{Croissant: A metadata format for ml-ready datasets}.
\newblock In \emph{\bibinfo{booktitle}{Proceedings of the Eighth Workshop on
  Data Management for End-to-End Machine Learning}}, \bibinfo{pages}{1--6}
  (\bibinfo{year}{2024}).

\bibitem{garbage}
\bibinfo{author}{Geiger, R.~S.} \emph{et~al.}
\newblock \bibinfo{title}{Garbage in, garbage out? do machine learning
  application papers in social computing report where human-labeled training
  data comes from?}
\newblock In \emph{\bibinfo{booktitle}{Proceedings of the 2020 Conference on
  Fairness, Accountability, and Transparency}}, FAT* '20,
  \bibinfo{pages}{325–336} (\bibinfo{publisher}{Association for Computing
  Machinery}, \bibinfo{address}{New York, NY, USA}, \bibinfo{year}{2020}).

\bibitem{healthcare}
\bibinfo{author}{Rostamzadeh, N.} \emph{et~al.}
\newblock \bibinfo{title}{Healthsheet: Development of a transparency artifact
  for health datasets}.
\newblock In \emph{\bibinfo{booktitle}{2022 ACM Conference on Fairness,
  Accountability, and Transparency}}, FAccT '22, \bibinfo{pages}{1943–1961}
  (\bibinfo{publisher}{Association for Computing Machinery},
  \bibinfo{address}{New York, NY, USA}, \bibinfo{year}{2022}).

\bibitem{diaz2022crowdworksheets}
\bibinfo{author}{D{\'\i}az, M.} \emph{et~al.}
\newblock \bibinfo{title}{Crowdworksheets: Accounting for individual and
  collective identities underlying crowdsourced dataset annotation}.
\newblock In \emph{\bibinfo{booktitle}{Proceedings of the 2022 ACM Conference
  on Fairness, Accountability, and Transparency}}, \bibinfo{pages}{2342--2351}
  (\bibinfo{year}{2022}).

\bibitem{emon2023mechanosensitive}
\bibinfo{author}{Emon, B.} \emph{et~al.}
\newblock \bibinfo{journal}{\bibinfo{title}{Mechanosensitive changes in the
  expression of genes in colorectal cancer-associated fibroblasts}}.
\newblock {\emph{\JournalTitle{Scientific Data}}}
  \textbf{\bibinfo{volume}{10}}, \bibinfo{pages}{350} (\bibinfo{year}{2023}).

\bibitem{ALDHABYANI2020104863}
\bibinfo{author}{Al-Dhabyani, W.}, \bibinfo{author}{Gomaa, M.},
  \bibinfo{author}{Khaled, H.} \& \bibinfo{author}{Fahmy, A.}
\newblock \bibinfo{journal}{\bibinfo{title}{Dataset of breast ultrasound
  images}}.
\newblock {\emph{\JournalTitle{Data in Brief}}} \textbf{\bibinfo{volume}{28}},
  \bibinfo{pages}{104863}, \url{https://doi.org/10.1016/j.dib.2019.104863}
  (\bibinfo{year}{2020}).

\bibitem{shen2018data}
\bibinfo{author}{Shen, Q.} \emph{et~al.}
\newblock \bibinfo{journal}{\bibinfo{title}{Data on the organic matter
  characteristics of new zealand soils under different land uses}}.
\newblock {\emph{\JournalTitle{Data in brief}}} \textbf{\bibinfo{volume}{21}},
  \bibinfo{pages}{620--638} (\bibinfo{year}{2018}).

\bibitem{xing2021optical}
\bibinfo{author}{Xing, Y.}, \bibinfo{author}{Duan, Y.},
  \bibinfo{author}{P.~Indurkar, P.}, \bibinfo{author}{Qiu, A.} \&
  \bibinfo{author}{Chen, N.}
\newblock \bibinfo{journal}{\bibinfo{title}{Optical breast atlas as a testbed
  for image reconstruction in optical mammography}}.
\newblock {\emph{\JournalTitle{Scientific Data}}} \textbf{\bibinfo{volume}{8}},
  \bibinfo{pages}{257} (\bibinfo{year}{2021}).

\bibitem{aerts2022pre}
\bibinfo{author}{Aerts, H.} \emph{et~al.}
\newblock \bibinfo{journal}{\bibinfo{title}{Pre-and post-surgery brain tumor
  multimodal magnetic resonance imaging data optimized for large scale
  computational modelling}}.
\newblock {\emph{\JournalTitle{Scientific Data}}} \textbf{\bibinfo{volume}{9}},
  \bibinfo{pages}{676} (\bibinfo{year}{2022}).

\bibitem{rodeles2016iberian}
\bibinfo{author}{Rodeles, A.~A.}, \bibinfo{author}{Galicia, D.} \&
  \bibinfo{author}{Miranda, R.}
\newblock \bibinfo{journal}{\bibinfo{title}{Iberian fish records in the
  vertebrate collection of the museum of zoology of the university of
  navarra}}.
\newblock {\emph{\JournalTitle{Scientific Data}}} \textbf{\bibinfo{volume}{3}},
  \bibinfo{pages}{1--7} (\bibinfo{year}{2016}).

\bibitem{cheng2022high}
\bibinfo{author}{Cheng, M.} \emph{et~al.}
\newblock \bibinfo{journal}{\bibinfo{title}{High-resolution crop yield and
  water productivity dataset generated using random forest and remote
  sensing}}.
\newblock {\emph{\JournalTitle{Scientific Data}}} \textbf{\bibinfo{volume}{9}},
  \bibinfo{pages}{641} (\bibinfo{year}{2022}).

\bibitem{brown2017standard}
\bibinfo{author}{Brown, A.~S.} \& \bibinfo{author}{Patel, C.~J.}
\newblock \bibinfo{journal}{\bibinfo{title}{A standard database for drug
  repositioning}}.
\newblock {\emph{\JournalTitle{Scientific data}}} \textbf{\bibinfo{volume}{4}},
  \bibinfo{pages}{1--7} (\bibinfo{year}{2017}).

\bibitem{steiger2018reconstruction}
\bibinfo{author}{Steiger, N.~J.}, \bibinfo{author}{Smerdon, J.~E.},
  \bibinfo{author}{Cook, E.~R.} \& \bibinfo{author}{Cook, B.~I.}
\newblock \bibinfo{journal}{\bibinfo{title}{A reconstruction of global
  hydroclimate and dynamical variables over the common era}}.
\newblock {\emph{\JournalTitle{Scientific data}}} \textbf{\bibinfo{volume}{5}},
  \bibinfo{pages}{1--15} (\bibinfo{year}{2018}).

\bibitem{lencioni2019human}
\bibinfo{author}{Lencioni, T.}, \bibinfo{author}{Carpinella, I.},
  \bibinfo{author}{Rabuffetti, M.}, \bibinfo{author}{Marzegan, A.} \&
  \bibinfo{author}{Ferrarin, M.}
\newblock \bibinfo{journal}{\bibinfo{title}{Human kinematic, kinetic and emg
  data during different walking and stair ascending and descending tasks}}.
\newblock {\emph{\JournalTitle{Scientific data}}} \textbf{\bibinfo{volume}{6}},
  \bibinfo{pages}{309} (\bibinfo{year}{2019}).

\bibitem{livneh2015spatially}
\bibinfo{author}{Livneh, B.} \emph{et~al.}
\newblock \bibinfo{journal}{\bibinfo{title}{A spatially comprehensive,
  hydrometeorological data set for mexico, the us, and southern canada
  1950--2013}}.
\newblock {\emph{\JournalTitle{Scientific data}}} \textbf{\bibinfo{volume}{2}},
  \bibinfo{pages}{1--12} (\bibinfo{year}{2015}).

\bibitem{funk2015centennial}
\bibinfo{author}{Funk, C.} \emph{et~al.}
\newblock \bibinfo{journal}{\bibinfo{title}{The centennial trends greater horn
  of africa precipitation dataset}}.
\newblock {\emph{\JournalTitle{Scientific data}}} \textbf{\bibinfo{volume}{2}},
  \bibinfo{pages}{1--17} (\bibinfo{year}{2015}).

\bibitem{venter2016sixteen}
\bibinfo{author}{Venter, O.} \emph{et~al.}
\newblock \bibinfo{journal}{\bibinfo{title}{Sixteen years of change in the
  global terrestrial human footprint and implications for biodiversity
  conservation}}.
\newblock {\emph{\JournalTitle{Nature communications}}}
  \textbf{\bibinfo{volume}{7}}, \bibinfo{pages}{12558} (\bibinfo{year}{2016}).

\bibitem{atzori2014electromyography}
\bibinfo{author}{Atzori, M.} \emph{et~al.}
\newblock \bibinfo{journal}{\bibinfo{title}{Electromyography data for
  non-invasive naturally-controlled robotic hand prostheses}}.
\newblock {\emph{\JournalTitle{Scientific data}}} \textbf{\bibinfo{volume}{1}},
  \bibinfo{pages}{1--13} (\bibinfo{year}{2014}).

\bibitem{tully2018reconstruction}
\bibinfo{author}{Tully, B.~J.}, \bibinfo{author}{Graham, E.~D.} \&
  \bibinfo{author}{Heidelberg, J.~F.}
\newblock \bibinfo{journal}{\bibinfo{title}{The reconstruction of 2,631 draft
  metagenome-assembled genomes from the global oceans}}.
\newblock {\emph{\JournalTitle{Scientific data}}} \textbf{\bibinfo{volume}{5}},
  \bibinfo{pages}{1--8} (\bibinfo{year}{2018}).

\bibitem{ta2018columbia}
\bibinfo{author}{Ta, C.~N.}, \bibinfo{author}{Dumontier, M.},
  \bibinfo{author}{Hripcsak, G.}, \bibinfo{author}{Tatonetti, N.~P.} \&
  \bibinfo{author}{Weng, C.}
\newblock \bibinfo{journal}{\bibinfo{title}{Columbia open health data, clinical
  concept prevalence and co-occurrence from electronic health records}}.
\newblock {\emph{\JournalTitle{Scientific data}}} \textbf{\bibinfo{volume}{5}},
  \bibinfo{pages}{1--17} (\bibinfo{year}{2018}).

\bibitem{brlik2021long}
\bibinfo{author}{Brl{\'\i}k, V.} \emph{et~al.}
\newblock \bibinfo{journal}{\bibinfo{title}{Long-term and large-scale
  multispecies dataset tracking population changes of common european breeding
  birds}}.
\newblock {\emph{\JournalTitle{Scientific data}}} \textbf{\bibinfo{volume}{8}},
  \bibinfo{pages}{21} (\bibinfo{year}{2021}).

\bibitem{tang2023chinese}
\bibinfo{author}{Tang, L.} \emph{et~al.}
\newblock \bibinfo{journal}{\bibinfo{title}{Chinese industrial air pollution
  emissions based on the continuous emission monitoring systems network}}.
\newblock {\emph{\JournalTitle{Scientific Data}}}
  \textbf{\bibinfo{volume}{10}}, \bibinfo{pages}{153} (\bibinfo{year}{2023}).

\bibitem{kohler2022meteorological}
\bibinfo{author}{Kohler, M.} \emph{et~al.}
\newblock \bibinfo{journal}{\bibinfo{title}{a meteorological dataset of the
  west african monsoon during the 2016 dacciwa campaign}}.
\newblock {\emph{\JournalTitle{Scientific Data}}} \textbf{\bibinfo{volume}{9}},
  \bibinfo{pages}{174} (\bibinfo{year}{2022}).

\bibitem{ajakaiye2018datasets}
\bibinfo{author}{Ajakaiye, O.~O.} \emph{et~al.}
\newblock \bibinfo{journal}{\bibinfo{title}{Datasets on factors influencing
  trading on pedestrian bridges along ikorodu road, lagos, nigeria}}.
\newblock {\emph{\JournalTitle{Data in brief}}} \textbf{\bibinfo{volume}{19}},
  \bibinfo{pages}{1584--1593} (\bibinfo{year}{2018}).

\bibitem{garcia2016time}
\bibinfo{author}{Garcia, D.}, \bibinfo{author}{Al~Nima, A.} \&
  \bibinfo{author}{Lindsk{\"a}r, E.}
\newblock \bibinfo{journal}{\bibinfo{title}{Time perspective and well-being:
  Swedish survey questionnaires and data}}.
\newblock {\emph{\JournalTitle{Data in Brief}}} \textbf{\bibinfo{volume}{9}},
  \bibinfo{pages}{183--193} (\bibinfo{year}{2016}).

\bibitem{lataifeh2020ar}
\bibinfo{author}{Lataifeh, M.} \& \bibinfo{author}{Elnagar, A.}
\newblock \bibinfo{journal}{\bibinfo{title}{Ar-dad: Arabic diversified audio
  dataset}}.
\newblock {\emph{\JournalTitle{Data in Brief}}} \textbf{\bibinfo{volume}{33}},
  \bibinfo{pages}{106503} (\bibinfo{year}{2020}).

\bibitem{paccotacya2022speech}
\bibinfo{author}{Paccotacya-Yanque, R.~Y.}, \bibinfo{author}{Huanca-Anquise,
  C.~A.}, \bibinfo{author}{Escalante-Calcina, J.},
  \bibinfo{author}{Ramos-Lov{\'o}n, W.~R.} \& \bibinfo{author}{Cuno-Parari,
  {\'A}.~E.}
\newblock \bibinfo{journal}{\bibinfo{title}{A speech corpus of quechua collao
  for automatic dimensional emotion recognition}}.
\newblock {\emph{\JournalTitle{Scientific Data}}} \textbf{\bibinfo{volume}{9}},
  \bibinfo{pages}{778} (\bibinfo{year}{2022}).

\bibitem{li2022data}
\bibinfo{author}{Li, K.} \& \bibinfo{author}{Jiao, C.}
\newblock \bibinfo{journal}{\bibinfo{title}{The data paper as a sociolinguistic
  epistemic object: A content analysis on the rhetorical moves used in data
  paper abstracts}}.
\newblock {\emph{\JournalTitle{Journal of the Association for Information
  Science and Technology}}} \textbf{\bibinfo{volume}{73}},
  \bibinfo{pages}{834--846} (\bibinfo{year}{2022}).

\bibitem{sdata}
\bibinfo{author}{Nature}.
\newblock \bibinfo{title}{Scientific data submissions guidelines}.
\newblock
  \bibinfo{howpublished}{\url{https://www.nature.com/sdata/publish/submission-guide}}
  (\bibinfo{year}{2023}).
\newblock \bibinfo{note}{Accessed: November 2023}.

\bibitem{dbrief}
\bibinfo{author}{Elsevier}.
\newblock \bibinfo{title}{Data in brief submissions guidelines}.
\newblock
  \bibinfo{howpublished}{\url{https://www.data-in-brief.com/content/authorinfo}}
  (\bibinfo{year}{2023}).
\newblock \bibinfo{note}{Accessed: November 2023}.

\bibitem{chen2023ai}
\bibinfo{author}{Chen, C.} \& \bibinfo{author}{Sundar, S.~S.}
\newblock \bibinfo{title}{Is this ai trained on credible data? the effects of
  labeling quality and performance bias on user trust}.
\newblock In \emph{\bibinfo{booktitle}{Proceedings of the 2023 CHI Conference
  on Human Factors in Computing Systems}}, \bibinfo{pages}{1--11}
  (\bibinfo{year}{2023}).

\bibitem{barbosa2019rehumanized}
\bibinfo{author}{Barbosa, N.~M.} \& \bibinfo{author}{Chen, M.}
\newblock \bibinfo{title}{Rehumanized crowdsourcing: A labeling framework
  addressing bias and ethics in machine learning}.
\newblock In \emph{\bibinfo{booktitle}{Proceedings of the 2019 CHI Conference
  on Human Factors in Computing Systems}}, \bibinfo{pages}{1--12}
  (\bibinfo{year}{2019}).

\bibitem{mahood2022country}
\bibinfo{author}{Mahood, A.~L.}, \bibinfo{author}{Lindrooth, E.~J.},
  \bibinfo{author}{Cook, M.~C.} \& \bibinfo{author}{Balch, J.~K.}
\newblock \bibinfo{journal}{\bibinfo{title}{Country-level fire perimeter
  datasets (2001--2021)}}.
\newblock {\emph{\JournalTitle{Scientific data}}} \textbf{\bibinfo{volume}{9}},
  \bibinfo{pages}{458} (\bibinfo{year}{2022}).

\bibitem{gebru2021datasheets}
\bibinfo{author}{Gebru, T.} \emph{et~al.}
\newblock \bibinfo{journal}{\bibinfo{title}{Datasheets for datasets}}.
\newblock {\emph{\JournalTitle{Communications of the ACM}}}
  \textbf{\bibinfo{volume}{64}}, \bibinfo{pages}{86--92}
  (\bibinfo{year}{2021}).

\bibitem{10.1145/3351095.3375709}
\bibinfo{author}{Yang, K.}, \bibinfo{author}{Qinami, K.},
  \bibinfo{author}{Fei-Fei, L.}, \bibinfo{author}{Deng, J.} \&
  \bibinfo{author}{Russakovsky, O.}
\newblock \bibinfo{title}{Towards fairer datasets: Filtering and balancing the
  distribution of the people subtree in the imagenet hierarchy}.
\newblock In \emph{\bibinfo{booktitle}{Proceedings of the 2020 Conference on
  Fairness, Accountability, and Transparency}}, FAT* '20,
  \bibinfo{pages}{547–558} (\bibinfo{publisher}{Association for Computing
  Machinery}, \bibinfo{address}{New York, NY, USA}, \bibinfo{year}{2020}).

\bibitem{benjamin2019towards}
\bibinfo{author}{Benjamin, M.} \emph{et~al.}
\newblock \bibinfo{journal}{\bibinfo{title}{Towards standardization of data
  licenses: The montreal data license}}.
\newblock {\emph{\JournalTitle{arXiv preprint arXiv:1903.12262}}}
  (\bibinfo{year}{2019}).

\bibitem{zhang2024missing}
\bibinfo{author}{Zhang, L.}, \bibinfo{author}{Cao, Z.}, \bibinfo{author}{Shang,
  Y.}, \bibinfo{author}{Sivertsen, G.} \& \bibinfo{author}{Huang, Y.}
\newblock \bibinfo{journal}{\bibinfo{title}{Missing institutions in openalex:
  possible reasons, implications, and solutions}}.
\newblock {\emph{\JournalTitle{Scientometrics}}} \bibinfo{pages}{1--23}
  (\bibinfo{year}{2024}).

\bibitem{brickley2019google}
\bibinfo{author}{Brickley, D.}, \bibinfo{author}{Burgess, M.} \&
  \bibinfo{author}{Noy, N.}
\newblock \bibinfo{title}{Google dataset search: Building a search engine for
  datasets in an open web ecosystem}.
\newblock In \emph{\bibinfo{booktitle}{The World Wide Web Conference}},
  \bibinfo{pages}{1365--1375} (\bibinfo{year}{2019}).

\bibitem{jiao2023exclusively}
\bibinfo{author}{Jiao, C.}, \bibinfo{author}{Li, K.} \& \bibinfo{author}{Fang,
  Z.}
\newblock \bibinfo{journal}{\bibinfo{title}{How are exclusively data journals
  indexed in major scholarly databases? an examination of four databases}}.
\newblock {\emph{\JournalTitle{Scientific Data}}}
  \textbf{\bibinfo{volume}{10}}, \bibinfo{pages}{737} (\bibinfo{year}{2023}).

\bibitem{candela2015data}
\bibinfo{author}{Candela, L.}, \bibinfo{author}{Castelli, D.},
  \bibinfo{author}{Manghi, P.} \& \bibinfo{author}{Tani, A.}
\newblock \bibinfo{journal}{\bibinfo{title}{Data journals: A survey}}.
\newblock {\emph{\JournalTitle{Journal of the Association for Information
  Science and Technology}}} \textbf{\bibinfo{volume}{66}},
  \bibinfo{pages}{1747--1762} (\bibinfo{year}{2015}).

\bibitem{walters2020data}
\bibinfo{author}{Walters, W.~H.}
\newblock \bibinfo{journal}{\bibinfo{title}{Data journals: incentivizing data
  access and documentation within the scholarly communication system.}}
\newblock {\emph{\JournalTitle{Insights: the UKSG journal}}}
  \textbf{\bibinfo{volume}{33}} (\bibinfo{year}{2020}).

\bibitem{priem2022openalex}
\bibinfo{author}{Priem, J.}, \bibinfo{author}{Piwowar, H.} \&
  \bibinfo{author}{Orr, R.}
\newblock \bibinfo{journal}{\bibinfo{title}{Openalex: A fully-open index of
  scholarly works, authors, venues, institutions, and concepts}}.
\newblock {\emph{\JournalTitle{arXiv preprint arXiv:2205.01833}}}
  (\bibinfo{year}{2022}).

\bibitem{scipdf}
\bibinfo{author}{Achakulvisut, T.}
\newblock \bibinfo{title}{\mbox{SciPDF} main repository}.
\newblock
  \bibinfo{howpublished}{\url{https://github.com/titipata/scipdf_parser}}
  (\bibinfo{year}{2023}).
\newblock \bibinfo{note}{Accessed: November 2023}.

\bibitem{romary:hal-01673305}
\bibinfo{author}{Romary, L.} \& \bibinfo{author}{Lopez, P.}
\newblock \bibinfo{journal}{\bibinfo{title}{{GROBID - Information Extraction
  from Scientific Publications}}}.
\newblock {\emph{\JournalTitle{{ERCIM News}}}} \textbf{\bibinfo{volume}{100}}
  (\bibinfo{year}{2015}).

\bibitem{tabula}
\bibinfo{author}{Ariga, A.}
\newblock \bibinfo{title}{Tabula main repository}.
\newblock \bibinfo{howpublished}{\url{https://github.com/chezou/tabula-py}}
  (\bibinfo{year}{2023}).
\newblock \bibinfo{note}{Accessed: November 2023}.

\bibitem{datadoc}
\bibinfo{author}{Giner-Miguelez, J.}, \bibinfo{author}{G{\'o}mez, A.} \&
  \bibinfo{author}{Cabot, J.}
\newblock \bibinfo{title}{Datadoc analyzer: A tool for analyzing the
  documentation of scientific datasets}.
\newblock In \emph{\bibinfo{booktitle}{Proceedings of the 32nd ACM
  International Conference on Information and Knowledge Management}},
  \bibinfo{pages}{5046--5050} (\bibinfo{year}{2023}).

\bibitem{lewis2020bart}
\bibinfo{author}{Lewis, M.} \emph{et~al.}
\newblock \bibinfo{title}{{BART}: Denoising sequence-to-sequence pre-training
  for natural language generation, translation, and comprehension}.
\newblock In \bibinfo{editor}{Jurafsky, D.}, \bibinfo{editor}{Chai, J.},
  \bibinfo{editor}{Schluter, N.} \& \bibinfo{editor}{Tetreault, J.} (eds.)
  \emph{\bibinfo{booktitle}{Proceedings of the 58th Annual Meeting of the
  Association for Computational Linguistics}}, \bibinfo{pages}{7871--7880}
  (\bibinfo{publisher}{Association for Computational Linguistics},
  \bibinfo{address}{Online}, \bibinfo{year}{2020}).

\bibitem{williams2018broad}
\bibinfo{author}{Williams, A.}, \bibinfo{author}{Nangia, N.} \&
  \bibinfo{author}{Bowman, S.}
\newblock \bibinfo{title}{A broad-coverage challenge corpus for sentence
  understanding through inference}.
\newblock In \bibinfo{editor}{Walker, M.}, \bibinfo{editor}{Ji, H.} \&
  \bibinfo{editor}{Stent, A.} (eds.) \emph{\bibinfo{booktitle}{Proceedings of
  the 2018 Conference of the North {A}merican Chapter of the Association for
  Computational Linguistics: Human Language Technologies, Volume 1 (Long
  Papers)}}, \bibinfo{pages}{1112--1122} (\bibinfo{publisher}{Association for
  Computational Linguistics}, \bibinfo{address}{New Orleans, Louisiana},
  \bibinfo{year}{2018}).

\bibitem{grootendorst2022bertopic}
\bibinfo{author}{Grootendorst, M.}
\newblock \bibinfo{journal}{\bibinfo{title}{Bertopic: Neural topic modeling
  with a class-based tf-idf procedure}}.
\newblock {\emph{\JournalTitle{arXiv preprint arXiv:2203.05794}}}
  (\bibinfo{year}{2022}).

\bibitem{giner2024using}
\bibinfo{author}{Giner-Miguelez, J.}, \bibinfo{author}{G{\'o}mez, A.} \&
  \bibinfo{author}{Cabot, J.}
\newblock \bibinfo{journal}{\bibinfo{title}{Using large language models to
  enrich the documentation of datasets for machine learning}}.
\newblock {\emph{\JournalTitle{arXiv preprint arXiv:2404.15320}}}
  (\bibinfo{year}{2024}).

\end{thebibliography}

\end{document}